%% file: main.tex
\documentclass[11pt, letterpaper]{article}
\usepackage[T1]{fontenc}
\usepackage[margin=1in]{geometry}

\usepackage{graphicx}
\usepackage{mathptmx}      %
\usepackage{flushend}
\usepackage[numbers,sort&compress]{natbib}
\usepackage[colorlinks,citecolor=blue,urlcolor=blue,linkcolor=blue]{hyperref}

\usepackage{xspace}
\usepackage{adjustbox}
\usepackage{booktabs}
\usepackage{import}
\usepackage{algorithm}
\usepackage{algorithmic}
\usepackage{amsmath}
\usepackage{amsfonts}
\usepackage{amssymb}
\usepackage{wrapfig}
\usepackage{subcaption}
\usepackage{multirow}
\usepackage{mdframed}
\usepackage{xcolor}

\subimport{./}{macro}

\graphicspath{ {./Figures/} }

\begin{document}

\title{
Model-Agnostic Utility-Preserving Biometric Information Anonymization
}

\author{Chun-Fu (Richard) Chen${}^{*1}$,
        Bill Moriarty${}^{*1}$, 
        Shaohan Hu${}^{*1}$, 
        Sean Moran${}^1$, \\
        Marco Pistoia${}^1$, 
        Vincenzo Piuri${}^2$, 
        Pierangela Samarati${}^2$ \\
${}^1$JPMorgan Chase Bank, N.A., USA\\
${}^2$Universit\`{a} degli Studi di Milano, Italy\\
\texttt{\{richard.cf.chen, william.r.moriarty, shaohan.hu\}@jpmchase.com}\\
\texttt{\{sean.j.moran, marco.pistoia\}@jpmchase.com}\\
\texttt{\{vincenzo.piuri, pierangela.samarati\}@unimi.it} \\
${}^*$Equal Contributions\\
}

\date{}
\vspace{-5pt}

\maketitle

\sloppy

\begin{abstract}
The recent rapid advancements in both sensing and machine learning technologies 
have given rise to the universal collection and utilization of people's biometrics,
such as fingerprints, voices, retina/facial scans, or gait/motion/gestures data,
enabling a wide range of applications including authentication, health monitoring,
or much more sophisticated analytics.
While providing 
better user experiences and deeper business insights, 
the use of biometrics has raised serious privacy concerns
due to their intrinsic sensitive nature 
and the accompanying high risk of leaking 
sensitive information such as identity or medical conditions.

In this paper, we propose 
a novel modality-agnostic data transformation framework
that is capable of anonymizing biometric data
by suppressing its sensitive attributes
and retaining features relevant to downstream
machine learning-based analyses that are of research and business values.
We carried out a thorough experimental evaluation using publicly available 
facial, voice, and motion datasets.
Results show that our proposed framework can achieve a \highlight{high suppression level
for sensitive information},
while at the same time retain underlying data utility such that
subsequent analyses on the anonymized biometric data
could still be carried out to yield satisfactory accuracy.
\end{abstract}

\input{./Tex/01_introduction}

\input{./Tex/03_method}

\input{./Tex/04_experiments}
\input{./Tex/02_related_works}

\input{./Tex/05_conclusions}

\section*{Acknowledgement}
Vincenzo Piuri and Pierangela Samarati were supported in part by the EC under project GLACIATION (101070141) and by project SERICS (PE00000014) under the NRRP MUR program funded by the EU - NGEU.

\section*{Disclaimer}
This paper was prepared for informational purposes with contributions from the Global Technology Applied Research center of JPMorgan Chase \& Co. This paper is not a product of the Research Department of JPMorgan Chase \& Co. or its affiliates. Neither JPMorgan Chase \& Co. nor any of its affiliates makes any explicit or implied representation or warranty and none of them accept any liability in connection with this paper, including, without limitation, with respect to the completeness, accuracy, or reliability of the information contained herein and the potential legal, compliance, tax, or accounting effects thereof. This document is not intended as investment research or investment advice, or as a recommendation, offer, or solicitation for the purchase or sale of any security, financial instrument, financial product or service, or to be used in any way for evaluating the merits of participating in any transaction.

\bibliographystyle{plain}
\bibliography{references}

\end{document}

%% file: macro.tex
\usepackage[normalem]{ulem}

\newcommand{\clustering}{{dynamically assembling a random set}\xspace}
\newcommand{\facegroup}{{dynamically assembled random set}\xspace}

\newif\ifhighlight

\highlighttrue
\highlightfalse %

\ifhighlight
\newcommand{\highlight}[1]{{\color{blue}{#1}}}
\newcommand{\highlightfig}[1]{
\begin{mdframed}[backgroundcolor=blue!100!white!100, linecolor=white] 
{#1}
\end{mdframed}}

\else
\newcommand{\highlight}[1]{#1}
\newcommand{\highlightfig}[1]{#1}

\fi

%% file: Tex/01_introduction.tex
\section{Introduction}\label{sec:intro}
As sensing technologies get increasingly adopted into commodity electronic devices that people use in their daily lives,
biometrics have become more accessible and appealing as an information source, 
for example to enable seamless authentication without manual password input~\cite{rui2018survey}.
What's more, the latest sensing technologies have gone way beyond just targeting more traditional biometrics such as fingerprints,
where the sole usage is arguably authentication only.
Today's sensing devices can collect rich biometrics such as
facial imagery, voice, and even posture/gait, iris, and neural signal data.
With the help of the recent rapid advances in machine learning techniques,
a wide range of interesting analytics can then be performed on the rich biometric data~\cite{ortiz2018survey},
for example, to infer or extract information such as age, gender, 
dialect, sentiment, emotion, focus level, medical condition, etc.,
which could then enable vast opportunities in various relevant services and business interests.

Despite the high potential value of biometric \highlight{data}, 
one major concern preventing its universal collection and utilization is its linkage to personal sensitive informaiton 
(e.g., identity, medical conditions, etc)
and the potential privacy violation~\cite{bdgps.access2021, datta2020survey, dpd.springer2013}.
For example, a user might enjoy the convenience of Face Unlock on their personal electronic devices,
but likely would not appreciate having their facial features and identity information collected
and used for targeted advertisements.
Similarly, businesses have deployed Voice ID authentication in their automated phone system
to streamline their customer service call experience.
\highlight{However}, it would be deeply problematic if a business extracts information such as age, gender, and race from the voice data
and uses it to profile each of their individual customers for preferential treatments.

\begin{figure}[t!]
    \centering
    \highlightfig{\includegraphics[width=0.5\linewidth]{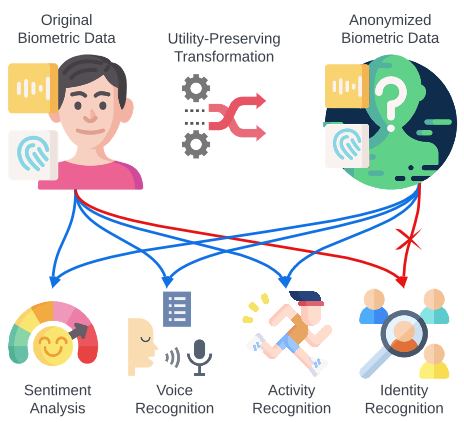}}
    \caption{\highlight{Our \textit{utility-preserving biometric information anonymization}
    transforms the original biometric data into an anonymized version
    such that the sensitive attributes can no longer be recognized from the transformed data records,
    but the rest of the attributes (which are useful and nonsensitive) can still be used for valuable analytics tasks.}}
    \label{fig:main}
\end{figure}

It is therefore our goal 
to devise a data transformation mechanism 
to resolve this conflict between the value of biometric data
and the potential 
\highlight{disclosure of sensitive information}.
\highlight{As illustrated in Fig.~\ref{fig:main}, 
the original data can be used for analytical tasks
such as sentiment analysis and activity recognition, 
but can also leak the sensitive identity information.
Our indented data transformation method would produce an anonymized version of the biometric data
such that the sensitive identity information can no longer be extracted,
but the rest of the nonsensitive, valuable attributes remain intact.}
Such a biometric anonymization mechanism would be tremendously valuable
across a multitude of use cases.
For example, a marketing firm that has recruited a focus group 
to study people's preference towards different products
by presenting to them series of images of new products 
and taking pictures of their facial reactions for analysis
might want to anonymize their collected facial imagery data
and transfer it to a technology company 
focusing on developing computer vision algorithms and software.
Or, an international medical research institute 
that has collected detailed biometric records from a large population 
might have completed their study of a particular disease
and would like to release an anonymized version of the dataset publicly
so other medical researchers could carry out their own studies on the dataset
and potentially make discoveries that are related,
or even orthogonal,
to the data's original purpose.

To make the data release and reuse possible,
the key challenge lies in the high dimensionality of biometric data
as well as in the intrinsic probabilistic nature of 
machine learning-based analytics performed on top of it.
In comparison,
for traditional tabular data
where the useful information associated with each data record
is simply the textual content itself
(e.g., date of birth, zip code, etc.),
a rich body of literature
exists that provides promising anonymization results.
For \highlight{biometrics}, on the other hand,
each data record on itself
(e.g., facial image, voice audio clip, etc.)
is essentially just a blob of bits,
and does not show its useful information without 
either manual labeling or 
automated machine learning-based analyses,
which by nature is probabilistic.
Even though from a philosophical point of view,
our goal of preserving interesting attributes and removing \highlight{sensitive ones}
might seem self-contradicting
in that any features preserved could potentially be used 
for \highlight{extracting the sensitive attributes},
we argue that our problem at hand around biometrics 
is far from being binary.
On the contrary, 
the high dimensionality of the data itself
and the probabilistic nature of machine learning-based analytics
introduce a high degree of uncertainty
that we can 
take advantage of
to achieve retention of interesting attributes
while performing anonymization.
\highlight{Our proposed method aims at achieving this very goal}.

The contribution of this paper is three-fold:
\begin{itemize}
    \item To the best of our knowledge, 
     \highlight{we propose a novel framework that}
     is the first one to introduce the concept of 
     \textit{utility preservation under the context of ML-based analytics with general biometric information anonymization}.
    
    \item \highlight{We introduce a novel technique 
    that uses a \facegroup 
    and task-oriented feature relevance metrics,
    either machine learning model-specific or purely data-driven,
    to guide a selective weighted-mean transformation
    for anonymizing or suppressing sensitive attributes from biometric data}.
    
    \item We demonstrate the effectiveness of our method's 
    \highlight{utility preservation and sensitive information suppressoin} via 
    a thorough experimental evaluation using publicly available 
    multi-modal datasets.
\end{itemize}

\section{Key Concepts}\label{sec:concepts}

Since our objective is to transform a private biometric dataset
for public release such that \highlight{sensitive information is suppressed}
but data utility is preserved as much as possible,
we would like to define a few terms we use 
for our proposed utility-preserving data anonymization task,
just so we are on level ground going forward with our discussion.

First, regarding the utility of a biometric dataset,
we define \textit{attribute of interest} and \textit{additional attributes},
as follows.
\begin{itemize}
    \item \textit{Attribute of Interest.}
An individual's biometric data contains features that can be used to 
predict certain attributes about them.
An {\em attribute of interest} is an 
attribute detectable from biometric data, whose value 
must be protected
\highlight{due to its potential research or business values}.
For example, the sentiment states displayed in a set of facial images
could be considered as an attribute of interest due to their potential
uses in computer vision studies or business applications.
\highlight{Hence in this case, when anonymizing such a facial image dataset,
we want to preserve the discoverability of sentiment states}.

    \item \textit{Additional Attribute.}
\highlight{We use the term \textit{additional attributes} to refer to
all other attributes (in addition to the \textit{attribute of interest})
that are detectable from the biometric data
and could potentially be of research or business values.}
For example, from a voice dataset,
information such as age group and dialect can be extracted
by analyzing audio clips.
If the age group information is the sole attribute of interest,
the dialect information \highlight{can be considered as} an additional attribute.
Preserving the dialect as well as the age group information
while anonymizing the voice dataset could be desirable 
for the expanded potential usages of a public release \highlight{of the dataset}.

\end{itemize}

\highlight{Additionally, we use the term \textit{Sensitive Attribute}
to refer to any attributes of the biometric data
that can be considered private or sensitive 
and should be removed (from the data) before the data can be safely released to the public.
For example, a sensitive attribute could be the identity information
recoverable from the biometric data.
It could also be race, gender, medical conditions,
or any other attribute of the biometrics 
that the data owner considers to be sensitive.
Therefore, in this paper when we talk about anonymization,
we are not only referring to the suppression of identity;
rather, any sensitive attribute could be the target.
}

\highlight{With the above definitions on the various \textit{attributes},
we also introduce two other notions that are key to our problem at hand,
as follows.
\begin{itemize}
    \item \highlight{
    \textit{Recognition Model.}
    We use the term \textit{recognition model} (or \textit{classification model}, or \textit{classifer})
    to refer to the machine learning algorithms that are used to extract/infer
    attribute values from the biometric data.
    For example, if we want to systematically obtain the sentiment information
    from a facial image dataset for legitimate business purposes,
    we could use a sentiment classifier that's already trained for the task.
    Similarly, an malicious actor might also attempt to use other classification models
    to try to extract the corresponding sensitive attributes, such as medical conditions.
    }
    
    \item \highlight{
    \textit{Data Transformation.}
    This is the core objective of this paper, that is,
    to compute a \textit{transformed} version of the original biometric data,
    such that the transformed data retains the utility of the original data
    but with all sensitive attributes removed.
    More specifically, the attribute of interest and the additional attributes
    should still be recognizable from the transformed data 
    via the corresponding recognition models,
    whereas the sensitive attributes should \textit{not}.
    } 
\end{itemize}
}

\section{\highlight{Problem Statement}}\label{sec:problem}

\highlight{With the various attributes, the recognition model, and the data transformation defined,
we next formulate our problem at hand and specify the particular type of attacks we consider from adversaries.}

\subsection{\highlight{Problem Formulation}}
\label{sec:problem-formulation}
Due to the high dimensionality of biometric data
and the high uncertainty of ML-based analytics,
we argue that it is impossible to formulate a provable security guarantee
for our biometric anonymization problem at hand.
Therefore,
we propose a purely \textit{data-driven} approach
so that the level of utility preservation
and the level of anonymization
can both be quantified, experimentally through measurements.

Consider an original biometric dataset $\mathbb{D}$,
and suppose that it has an attribute of interest $p$ 
and a set of $n$ additional attributes $\{q_n\}$,
with their corresponding recognition models
$\mathnormal{P}(\cdot)$ and $\{\mathnormal{Q}_n(\cdot)\}$
all trained from the original dataset $\mathbb{D}$.
Suppose $\mathbb{D}$ has 
\highlight{a sensitive attribute}
classification model $\mathnormal{S}(\cdot)$, 
also trained from the original dataset.
Then, for any data transformation $\mathnormal{T}(\cdot)$,
\highlight{we can define the
\textit{Utility} $U(\cdot)$ of the transformed data
with respect to the attribute of interest and additional attributes
as their collective recognition accuracy}
\[
U(\mathnormal{T}(\mathbb{D}))=\mathnormal{P}(\mathnormal{T}(\mathbb{D})) + \sum_{i=1}^n \alpha_i \mathnormal{Q}_i(\mathnormal{T}(\mathbb{D})),
\]
and
what we call 
\textit{Mixture} $M(\cdot)$
the degree to which the trained \highlight{sensitive attribute} classification model
is confused by the transformed data
\[
M(\mathnormal{T}(\mathbb{D}))=1-\mathnormal{S}(\mathnormal{T}(\mathbb{D})).
\]
In the formulae,
$\mathnormal{T}(\mathbb{D})$ is the transformed biometric dataset, $\{\alpha_n\}$ are user input weights 
for the additional attributes.
Each of the attribute recognition models
$\mathnormal{P}(\cdot)$ and $\{\mathnormal{Q}_n(\cdot)\}$,
as well as the
\highlight{sensitive attribute}
classification model $\mathnormal{S}(\cdot)$,
takes as input an entire dataset
and outputs its accuracy.
Intuitively, to find the best anonymization 
for a biometric dataset $\mathbb{D}$
is to find the optimal $\mathnormal{T}^*(\cdot)$
that maximizes both $U$ and $M$ 
(or achieves a good trade-off between them),
which is to say that the corresponding transformed data
thoroughly confuses the \highlight{sensitive attribute} classification model
but can still be used to reliably extract interesting attributes.

\subsection{\highlight{Adversary} Model}\label{sec:attack-model}

\highlight{Given our problem formulation}, 
we would like to make an important observation
on the \highlight{sensitive attribute} classification model $\mathnormal{S}(\cdot)$:
Only the data owner knows the ground-truth for the
sensitive attributes (e.g., the true identity correspondence
between the original data $\mathbb{D}$
and the transformed data $\mathnormal{T}(\mathbb{D})$).
Therefore, only the data owner can compute the accuracy
$\mathnormal{S}(\mathnormal{T}(\mathbb{D}))$.
Any attacker who tries to use a 
\highlight{sensitive attribute
classifier
$\mathnormal{S}'(\cdot)$
would not be able to have any certainty in the obtained result $\mathnormal{S}'(\mathnormal{T}(\mathbb{D}))$
due to the \highlight{probablistic nature of machine learning-based analytics}.}
Therefore, even if the attacker's model $\mathnormal{S}'(\cdot)$
correctly classified $x\%$ 
of the hidden \highlight{sensitive attributes}
in $\mathnormal{T}(\mathbb{D})$,
the attacker would not be able to tell which $x\%$
in $\mathnormal{T}(\mathbb{D})$ 
the model $\mathnormal{S}'(\cdot)$ got correctly.
Thus, their attempted attack is
reduced to random guess.

Additionally, we argue that 
it is reasonable to assume the data owner's model 
is always more powerful than the attacker's model, 
\highlight{formally,}
$\forall{\mathbb{D}}:\mathnormal{S}(\mathbb{D}) \ge \mathnormal{S}'(\mathbb{D})$.
\highlight{In fact,} the data owner and the attacker 
can both select the latest and most powerful classification algorithm,
but the data owner has the advantage 
of having access to the original unanonymized data,
which the attacker does not have.
Therefore, if we let the attacker's model be the same as its upper bound,
$\mathnormal{S}'(\cdot) = \mathnormal{S}(\cdot)$,
we can treat the data owner's measured mixture $M$
to be the lower bound of what the attacker can possibly experience.
In other words, \textit{the already hidden \highlight{sensitive attribute} in the anonymized data
would appear even more mixed to an attacker}.
Therefore, in our discussion,
we assume that
{\em i)} the data owner only releases the final anonymized data, 
and nothing else,
and
{\em ii)}
the sensitive attribute classification model used by the attacker 
is effectively the same
as the model used by the data owner.

Our adversary model gives us a solid ground for 
our subsequent discussions.
We believe that, in practice, 
our data-driven approach can bring value to 
a wide range of application scenarios.

%% file: Tex/03_method.tex
\section{Methodology}
\label{sec:method}

In this section, we discuss our proposed framework
for performing utility-preserving anonymization on biometric data.
Our proposal is generally applicable to all types of biometrics,
and not restricted to any particular data modalities,
feature extraction methods,
\highlight{or targeted sensitive attributes,
as experimentally demonstrated in Sec.~\ref{sec:eval}.}

\subsection{Rationale of \highlight{Our} Approach}\label{sec:rationale}

To achieve our objective of utility-preserving anonymization for biometrics,
the high dimensionality of the data
and the uncertainty of ML-based analytics
need to be accounted for.
For each data record $\mathbf{d}$ we aim to anonymize,
we dynamically assemble a random set containing $\mathbf{d}$
and perform a selective weighted-mean-based operation,
where the weighting is only applied to the most important features,
as guided by task-specific machine learning models.
We intend to make our data transformation 
retain as much truthfulness as possible,
hence our approach follows the intuition of
only utilizing information from the original biometric dataset,
and purposefully avoiding external artificial noise.
Therefore, the transformation step $\mathnormal{T}(\cdot)$
randomly assembles a short-lived, 
parameter-driven 
(such parameters include desired set size, attribute purity, etc.,
which are discussed in detail in Sec.~\ref{sec:method:clustering})
set of feature vectors with which to calculate
the weighted-mean for each of the target feature vectors being anonymized.

Under our proposal,
each data record becomes different from its original form. 
Also, due to the high dimensional nature of biometrics,
it is also \highlight{highly} unlikely
for any anonymized data record to have an exact match
in the original dataset,
or vice versa.
As will be demonstrated in Sec.~\ref{sec:eval:results:quality},
regardless of the particular attack method of choice---be 
it a direct distance measure between two data records
or via a trained ML model to compute the probability of a 
match---the likelihood of an attacker being able to link any
anonymized data record to its true corresponding original record
is reduced to a random guess
on the entire dataset.
In other words, an anonymized record is equally likely
to be the closest, or the farthest, or anywhere in between,
to its true match,
as far as re-identification is concerned.
\highlight{Similarly, as will be experimentally demonstrated in Sec.~\ref{subsubsec:other_sensitive},
attack attempts on recovering other sensitive attributes would also be ineffective
on the transformed data.
}

\subsection{Dynamically Assembled Random Set}\label{sec:method:clustering}

Regardless of the particular preprocessing and feature extraction,
each biometric data record $\mathbf{d}$ is essentially a feature vector,
which we transform through a series of operations,
starting with \clustering of other data records
from the dataset to be anonymized.

For each target data record $\mathbf{d}$ 
in the original dataset $\mathbb{D}$,
we assemble a set $G$ of $g=|G|$ data records 
where $\mathbf{d} \in G$,
and the rest $g-1$ of the data records, $G\setminus \{\mathbf{d}\}$,
are selected from $\mathbb{D}$ based on their
attribute-of-interest values,
according to a \textit{purity} parameter
\[
t=\frac{|\{\mathbf{g}\in G | p_\mathbf{g} = p_\mathbf{d}\}|}{g},
\]
where $p_\mathbf{g}$ denotes
the value of the $\mathbf{g}$'s attribute of interest.
For example, if $t=1$, then all $\mathbf{g}$ in $G$
share the same attribute-of-interest value as $\mathbf{d}$;
if $t=\frac{|\{\mathbf{k}\in \mathbb{D} | p_\mathbf{k} = p_\mathbf{d}\}|}{|\mathbb{D}|}$,
which is the proportion of $p_\mathbf{d}$ in the entire population $\mathbb{D}$,
then all of $G$'s elements
are to be uniformly randomly selected from $\mathbb{D}$ 
regardless of their attribute-of-interest values;
if $t=\frac{1}{g}$, then all other elements in $G$
are selected to be of different attribute-of-interest values than $\mathbf{d}$.
This way of assembling the set $G$ is inspired by the
$k$-anonymity, $\ell$-diversity, and $t$-closeness methods,
but differs in that our approach was designed
with the main objective of preserving the attribute of interest,
while also including mechanisms for trading off between
\highlight{preserving the attribute of interest and suppressing sensitive information},
in the form of different set sizes $g \in \mathbb{Z}^+$
and purity levels $t \in \left[\frac{1}{g}, 1\right]$.
\highlight{Another key distinction is that
our proposed approach does not produce groups of identical data records 
as the aforementioned existing methods do;
rather, all the produced records will be different from each other
as well as from their own original values.
}

\subsection{Selective Weighted Mean-based Transformation}
\label{sec:method:weighted-mean}
\highlight{After \clustering $G$ around the target $\mathbf{d}$, 
we transform $\mathbf{d}$ by computing its \textit{weighted} mean 
with the rest of $G$'s elements $G \setminus \{\mathbf{d}\}$,
where the higher weights for $\mathbf{d}$ help prevent
its features from getting completely buried 
when it is averaged with the rest of $G$'s elements.
However, instead of protecting all the features of $\mathbf{d}$,
we want to protect only the subset of features that are relevant to the interesting downstream tasks,
and, if possible, exclude those that can be used to identify $\mathbf{d}$ or infer its senstive attributes.
Therefore, we want to make sure that the weights are only applied to selective 
features that meet the above criteria.
For example,
suppose we want to remove the identity information from a facial-image dataset
and preserve the sentiment information,
where the sentiment state of a face can be estimated from the eyes and the mouth,
and the identity of a face can be determined by the eyes and the nose,
then we might consider the following strategy:
\begin{enumerate}
    \item We definitely want to protect the mouth features because they are relevant to the sentiment and cannot be used for reidentification;
    \item We definitely want to exclude the nose features because they are not even relevant to the sentiment and, to make matters worse, they can be used for reidentification; and
    \item We might or might not want to protect the eyes features, because even though they are relevant to the sentiment, they might increase the risk of reidentification.
\end{enumerate}

The natural question to ask next would be
how to quantify a feature's relevance to any particular task.
There are multiple different routes.
For example, certain classification algorithms already provide such metrics,
like the Gini importance scores from decision tree-based methods.
Alternatively, we can compute certain statistical measures,
such as the mutual information between a feature and an attribute from the dataset ifself, 
independent from any particular classification methods.
With the feature relevance scores,
we can then proceed with the selective weighting
by ranking all the features
and prioritize the weighting for features 
that have higher relevance scores for the attribute of interest
and lower relevance scores for \highlight{the sensitive attributes}.
Also, note that not only can the attribute of interest
be preserved by the selective weighting,
so can any additional attributes.
For example,
in addition to sentiment,
the data owner of a facial image dataset
might also want to preserve hair style as an additional attribute,
in which case hair-related features in the facial images will most likely have high relevance scores
and be selected to receive the higher weights}.

\begin{algorithm}[t]\small
\caption{Model-Agonstic Utility-Preserving Biometric Information Anonymization}
\label{alg:alg-sampling}
    $\begin{array}{rl}
    \textbf{Inputs:} &\\
        \highlight{\mathbb{D}}: & \highlight{\text{the set of original biometric dataset to be anonymized}}\\
        \highlight{p}: & \highlight{\text{the attribute of interest}}\\
        \mathnormal{P(\cdot)}: & \text{the pretrained classifier for the attribute of interest}\\
        \highlight{\{q_n\}}: & \highlight{\text{the set of $n$ additional attributes}}\\
        \{\mathnormal{Q}_n(\cdot)\}: & \text{the pretrained classifiers for the additional attributes}\\
        \highlight{s}: & \highlight{\text{the sensitive attribute}}\\
        \highlight{\mathnormal{S(\cdot)}}: & \highlight{\text{the pretrained classifier for the sensitive attribute}}\\
        \highlight{\mathnormal{F(\cdot)}}: & \highlight{\text{the feature selection function based on relevance scores}}\\
        g: & \text{the size of random set for anonymization}\\
        t: & \text{the purity of the random set's attribute-of-interest value}\\
        w: & \text{the weight parameter for computing weighted mean }\\
    \textbf{Output:} & \\
        \mathbb{D}': & \text{the set of anonymized biometric data feature vectors}
    \end{array}$
    \vspace{.1cm}
    \hrule
    \vspace{.1cm}
  \begin{algorithmic}[1]
    \STATE \highlight{$R_p \leftarrow$ the feature relevance scores for the attribute of interest,
            obtained either directly from the pretrained classification model $\mathnormal{P(\cdot)}$
            or by computing data-driven statistics such as mutual information on $D$ and $p$}
    \STATE \highlight{$\{R_{q_n}\} \leftarrow$ the set of feature relevance scores for the additional attributes,
            obtained similarly as $R_p$}
    \STATE \highlight{$R_s \leftarrow$ the feature relevance scores for the sensitive attribute,
            obtained similarly as $R_p$}
    \STATE \highlight{$I \leftarrow \mathnormal{F}(R_p, \{R_{q_n}\}, R_s)$, the set of feature indices selected by $\mathnormal{F}$}
    \STATE $X_I \leftarrow$ indicator vector s.t.
            $X_I[j] = \left\{\begin{array}{lr}
                1, & \text{if } j \in I \\
                0, & \text{o.w.}
                \end{array}\right.$

    \STATE $\mathbb{D}' \leftarrow \emptyset$
    \FOR{\textbf{each} $\mathbf{d} \in \mathbb{D}$}
        \STATE Randomly select $G \subseteq \mathbb{D}$ s.t. $\mathbf{d} \in G$, $|G| = g$, and $\frac{|\{\mathbf{g}\in G | p_\mathbf{g} = p_\mathbf{d}\}|}{g}=t$
        \STATE \highlight{$\mathbf{d}' \leftarrow 
            \frac{(w-1)\cdot\mathbf{d} + \sum_{\mathbf{g} \in G}{\mathbf{g}}}{(w-1) + g} \odot X_I +
            \frac{\sum_{\mathbf{g} \in G}{\mathbf{g}}}{g} \odot (\mathbf{1} - X_I)$}
        \STATE $\mathbb{D}' \leftarrow \mathbb{D}' \cup \{\mathbf{d}'\}$
    \ENDFOR
    
    \RETURN $\mathbb{D}'$
  \end{algorithmic}
\end{algorithm}

\highlight{Algorithm~\ref{alg:alg-sampling} privides the pseudo-code of our approach.
Line 1 through 3 prepare the feature relevance scores
for the attribute of interest, 
as well as the additional attributes 
and the sensitve attribute.
This can be done either by directly accessing such feature scores from the classification model itself (e.g. Gini score from decision tree-based algorithms),
or by computing model-agnostic statistical metrics
such as mutual information between each feature and each attribute.
Taking into consideration all these feature relevance scores,
Line 4 then selects a subset of features
to be included for the upcoming selective weighting,
where the corresponding indicator vector for the selected features
is prepared in Line 5.}
Line \highlight{8} prepares the \facegroup as discussed in 
Sec.~\ref{sec:method:clustering}.
The selective weighted mean
as discussed in Sec.~\ref{sec:method:weighted-mean}
is computed on Line \highlight{9}, 
where $\odot$ is the component-wise multiplication operator.
Note that even though we only use a single weight $w$ here,
the algorithm can be easily extended 
to incorporate multiple weights, one per attribute for example,
by modifying the indicator-vector preparation on Line \highlight{5}
and/or the averaging computation on Line \highlight{9}.
Lastly, each iteration of the for-loop in Line \highlight{7} through \highlight{11}
is independent from the rest,
leading to highly parallelizable
and efficient computation in practice.

%% file: Tex/04_experiments.tex
\section{Experimental Evaluation}\label{sec:eval}
In this section, we experimentally evaluate our biometric anonymization technique 
using publicly available datasets. 
First, we describe the characteristics of the datasets we use,
and the experimental settings.
Next, we report the results of the various sets of experiments 
where we compare the effects of parameters in our proposed technique,
examining its capabilities for mixture 
and biometric attribute preservation under various experimental settings.

\subsection{Experimental Setup}\label{sec:eval:setup}

\highlight{We next introduce the three datasets used in our experiments,
our evaluation protocol,
the exact feature ranking and selection scheme
(i.e., $\mathnormal{F}$ from Alg.~\ref{alg:alg-sampling}),
and our experimental parameter settings.}

\subsubsection{Datasets}\label{sec:eval:setup:datasets}

Our framework enables the preservation of multiple attributes of biometric data 
while performing anonymization. 
Thus, an ideal dataset for us to use to demonstrate this capability would be one 
that contains ground-truth label information for multiple interesting attributes. 
We curated \highlight{three} publicly available datasets 
that fitted our requirement for testing our \highlight{proposed} method.

The first one was the facial image 
FER-2013 dataset~\cite{goodfellow2013challenges},
which contains grayscale images of human faces with associated ground-truth sentiment label information, and thus suits our purpose. 
We performed a round of manual inspection on the original dataset to remove problematic images that were duplicates, 
non-photographic, or of poor resolution, etc. 
We treated {\em sentiment} as an example biometric attribute of interest in our experiments \highlight{while taking the {\em identity} of each image as the sensitive attribute.}
Moreover, we augmented FER-2013 with 
the {\em mouth-slightly-open} attribute 
using a model pre-trained on the CelebFaces Attributes 
(CelebA) dataset~\cite{liu2015faceattributes}
as an additional attribute.
As a result, the final in-use FER-2013 dataset had 8,470 training images, 978 validation images and 1,060 testing images, and it had 4 classes for the sentiment attribute and two classes for the mouth-slightly-open attribute.

The \highlight{second} one was the voice AudioMNIST~\cite{audiomnist} dataset, 
which contains the voice audio clips of \highlight{24 different people speaking the 10 different digits}.
\highlight{We selected two sensitive attributes of each speaker, namely \textit{idenitiy} and \textit{age}.
For primary results,
we used {\em spoken digit} as the attribute of interest in our experiments and {\em speaker identity} as the sensitive attribute,
and we further tested using {\em age} as senstive attribute in Sec.~\ref{subsubsec:other_sensitive}.}
We \highlight{sub-sampled} the dataset to rebalance the different classes
since the original class \highlight{distribution was} highly skewed. 
We ended up with 7,200 training \highlight{samples}, 
2,400 validation \highlight{samples},
and 2,400 testing \highlight{samples}.

\highlight{The third one was the MotionSense dataset~\cite{Malekzadeh:2019:MSD:3302505.3310068}, a smartphone motion sensor dataset collected from 24 people as they performed 4 different physical activities, namely going up/downstairs, walking, and jogging.
This dataset also includes two attributes that we could consider sensitive: people's identity and gender information.
In our experiments, 
we treated \textit{activity} as the attribute of interest,
and \textit{identity} as the targeted sensitive attribute to be suppressed.
We also further validated our method using {\em gender} as the sensitive attribute, as discussed in detail in Sec.~\ref{subsubsec:other_sensitive}.
We splitted the dataset into 60,980 training and 13,344 validation samples. 

For the FER-2013 and AudioMNIST datasets, 
their existing training and validation splits were used to train the classifiers whereas the testing split was used to evaluate our proposed method. 
For the MotionSense dataset, due to its lack of a testing split, we trained on the training split and evaluated on the validation split.}

\subsubsection{Data Preprocessing} For FER-2013, 
we experimented with multiple feature extraction methods 
as the representation for each data records:
{\em i)} FaceGraph, the fully-connected graph built on facial landmarks
extracted from each facial image 
(using the Swift Vision Library~\cite{swift-vision});
{\em ii)} Pixel, the raw pixel values of an image
as the feature vector;
{\em iii)} Eigenface~\cite{newton2005preserving}, the projection
of an facial image onto the eigenspace computed from all facial images; and
{\em iv)} Vggfeats, the feature 
of the final layer of the facenet~\cite{FaceNet} neural network.
For AudioMNIST, on the other hand, 
we \highlight{extracted} the embeddings by using HuBERT-L~\cite{Hubert} on the voice signal 
and then \highlight{averaged} the embeddings of each token 
as our final data representation.
\highlight{For MotionSense, we followed the preprocessing steps in the original paper~\cite{Malekzadeh:2019:MSD:3302505.3310068}, which concatenated the raw sensor signals of rotationRate and userAcceleration as the data representation.}

\subsubsection{Evaluation Protocol}
We \highlight{used} the {\em classification on attribute of interest} 
as a driving example for our experiments.
Each classification task itself, however,
\highlight{was} {\em not} necessarily our focus.
Our goal \highlight{was} not 
to find the model that achieves absolutely the best accuracy for
a classification test.
Rather, we \highlight{were} mostly interested in
demonstrating
that a model trained on the original biometric data
\highlight{could} continue to successfully perform classification tasks
even on the version of the biometric data
transformed by our anonymization method.
Therefore, we experimented with 
a few well-known classification algorithms
and empirically picked the one that struck a balance 
between classification performance and training speed.
We finally picked the off-the-shelf 
Random-Forest~\cite{breiman2001random} classifier 
from scikit-learn~\cite{scikit-learn} 
for our experiments for attribute classification:
It provided good accuracy on both the attribute of interest 
and the additional attribute on \highlight{all three datasets}.

The evaluation protocol \highlight{was} setup as follows. 
First, to evaluate the preserved attribute of interest, 
we \highlight{trained} and \highlight{tested} a random-forest classifier on the original unanonymized data.
We then \highlight{applied} this classifier
on the anonymized data to check the level of preservation
on the attribute of interest. 
We also \highlight{evaluated} the \highlight{level of mixture} on the anonymized data.
Since the FER-2013 dataset does not \highlight{contain} identity information, 
when evaluating the \highlight{level of mixture}, 
\highlight{we considered each image to be of its own different identity} 
and then \highlight{measured} the {\em cosine distance} between 
each anonymized data record to all originals to find the closest one 
as the potential match.
AudioMNIST \highlight{and MotionSense}, on the other hand, do include the identity information.
So, we \highlight{employed an ML-based method to measure the level of mixture},
where we \highlight{trained} a multi-layer perceptron (MLP) over the identity labels 
using the original dataset and then \highlight{evaluated} its performance 
on the anonymized dataset \highlight{for AudioMNIST while using random-forest for MotionSense}.

\subsubsection{Feature Ranking and Selection}
\highlight{
As discussed in Sec.~\ref{sec:method:weighted-mean},
our method involves ranking and selecting subset of the features 
based on their relevance scores estimated 
either from the specific classification models 
(e.g., the feature importance scores computed by the random-forest classifier)
or by computing statistical measures such as mutual information
on the dataset inrespective of the classification model in use,
or via other existing feature selection methods~\cite{InfiniteFeatureSelection, FisherFeatureSelection}.
}
\highlight{
For the majority of our experiments,
we used the random-forest classifier to estimate feature relevance scores.
We also validated our method in by using mutual information as the feature relevance estimator in Sec.~\ref{sec:eval:diff-relevance-estimator}.
After obtaining the feature relevance scores for the attribute of interest and the additional attributes, 
we ordered the scores for each attribute and selected a fixed ratio (i.e., feature retention ratio).
Additionally, Sec.~\ref{subsubsec:suppress}
provides more details on the usage of feature relevance scores for the sensitive attributes.
}

\subsubsection{Parameter Settings} \label{subsubsec:parameters}
\begin{figure*}[t!]
    \centering
    \begin{subfigure}[t]{0.31\linewidth}
        \centering
        \includegraphics[width=\linewidth]{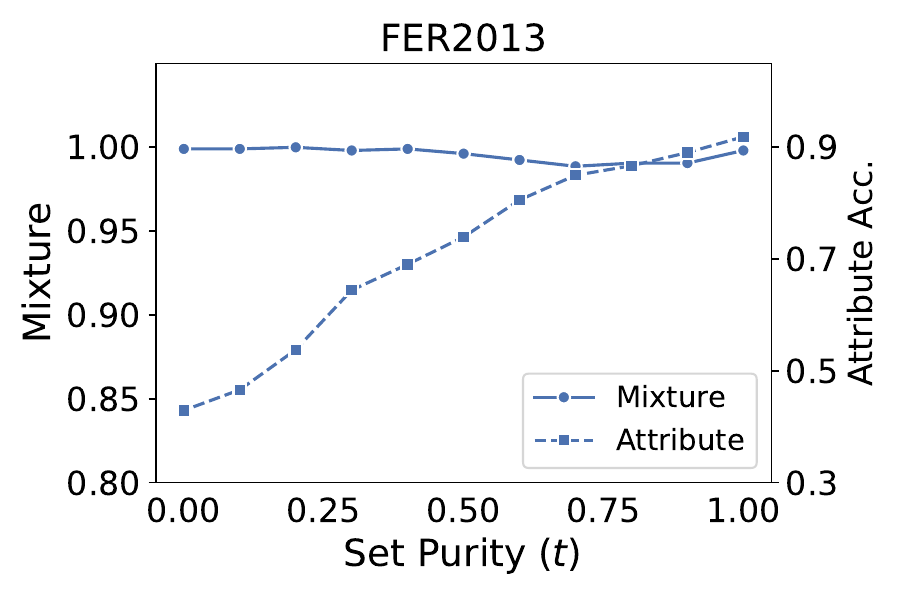}
    \caption{\footnotesize $g=32, w=100, r_p=1\%, r_q=0\%$}
    \end{subfigure}%
    ~ 
    \begin{subfigure}[t]{0.31\linewidth}
        \centering
        \includegraphics[width=\linewidth]{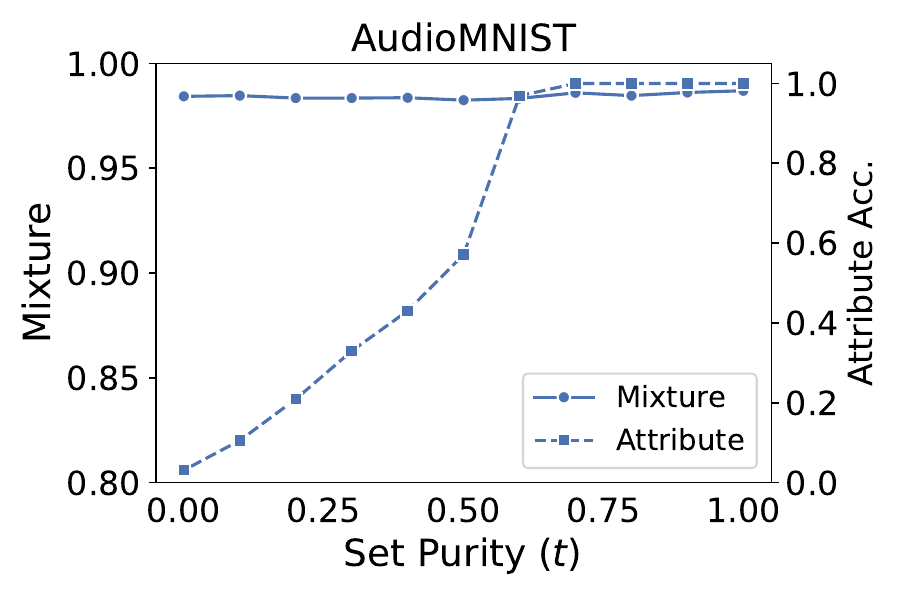}
    \caption{\footnotesize $g=128, w=10, r_p=1\%, r_q=0\%$}
    \end{subfigure}
    ~ 
    \begin{subfigure}[t]{0.31\linewidth}
        \centering
        \includegraphics[width=\linewidth]{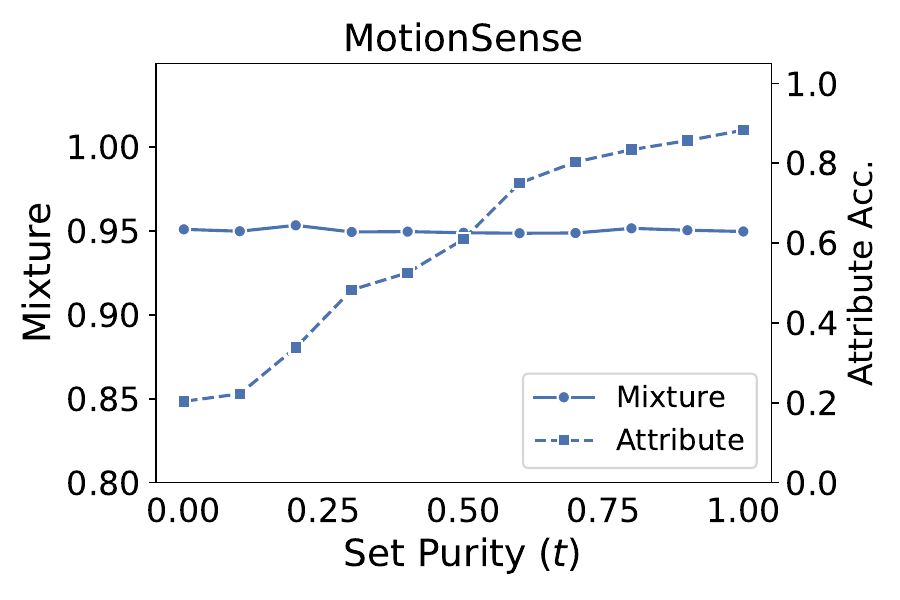}
    \caption{\highlight{\footnotesize $g=32, w=100, r_p=10\%, r_q=0\%$}}
    \end{subfigure}
    \caption{Varying set purity $t$. Higher $t$ leads to better attribute-of-interest
            recognition accuracy as each original data record $\mathbf{d}$ is combined 
            with more records sharing $\mathbf{d}$'s attribute value.}
    \label{fig:abl_t}
\end{figure*}

\begin{figure*}[t!]
    \centering
    \begin{subfigure}[t]{0.31\linewidth}
        \centering
        \includegraphics[width=\linewidth]{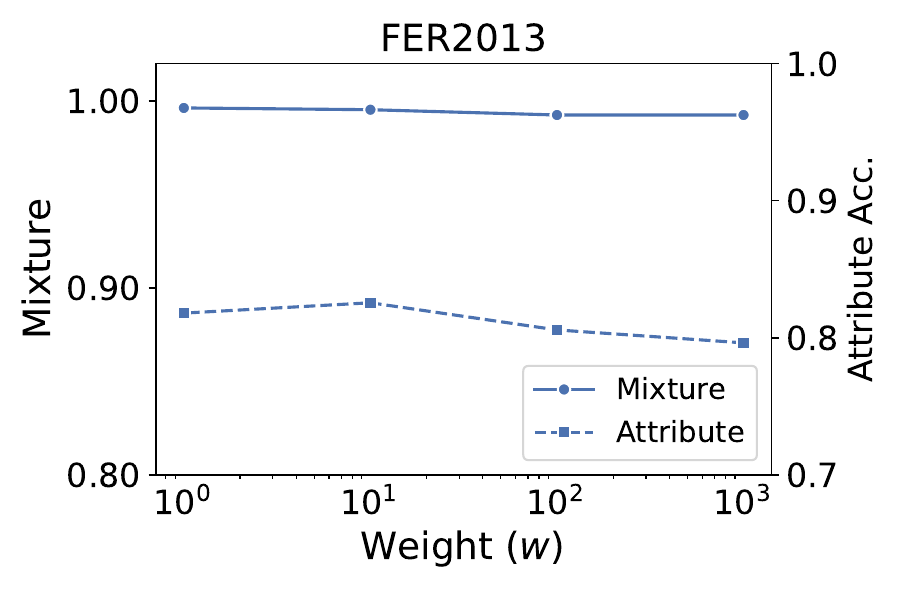}
    \caption{\footnotesize $g=32, t=0.6, r_p=1\%, r_q=0\%$}
    \end{subfigure}%
    \begin{subfigure}[t]{0.31\linewidth}
        \centering
        \includegraphics[width=\linewidth]{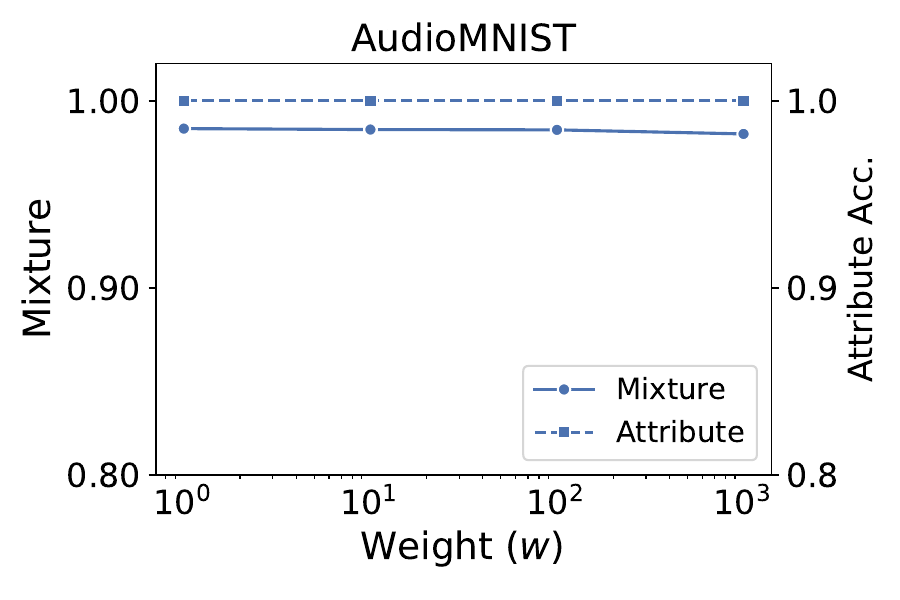}
    \caption{\footnotesize $g=128, t=0.8, r_p=1\%, r_q=0\%$}
    \end{subfigure}
    \begin{subfigure}[t]{0.31\linewidth}
        \centering
        \highlightfig{
        \includegraphics[width=\linewidth]{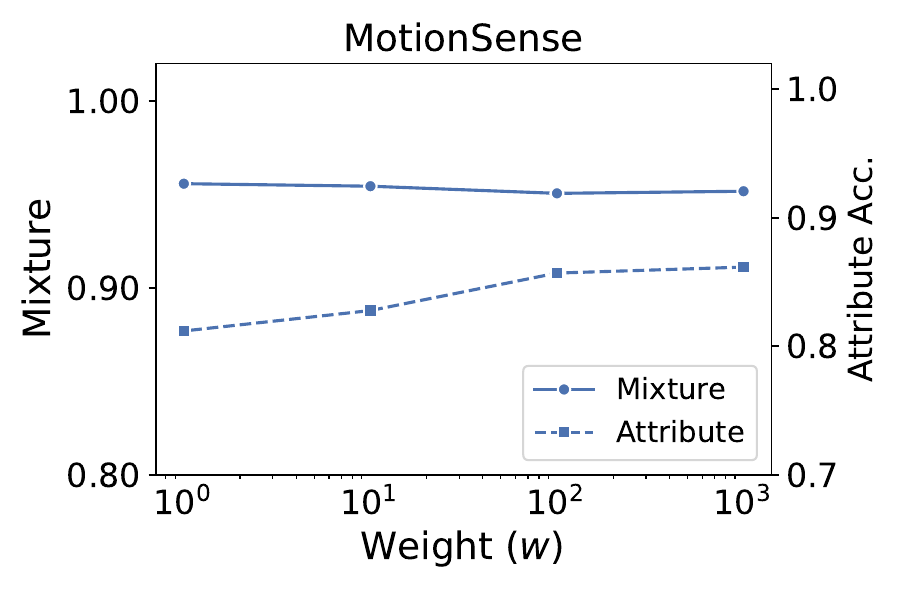}}
    \caption{\highlight{\footnotesize $g=32, t=0.9, r_p=10\%, r_q=0\%$}}
    \end{subfigure}
    \\
    \begin{subfigure}[t]{0.31\linewidth}
        \centering
        \includegraphics[width=\linewidth]{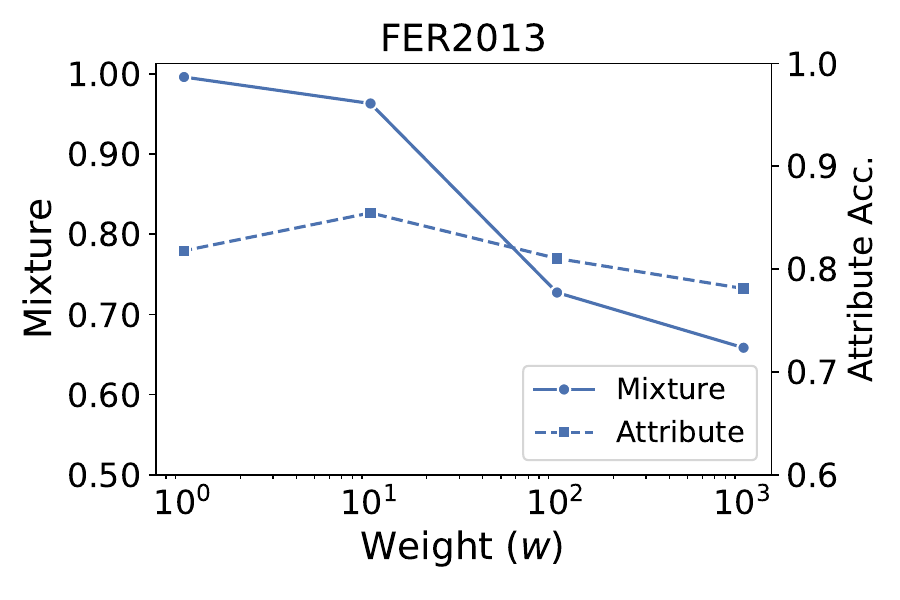}
    \caption{\footnotesize $g=32, t=0.6, r_p=50\%, r_q=0\%$}
    \end{subfigure}%
    \begin{subfigure}[t]{0.31\linewidth}
        \centering
        \includegraphics[width=\linewidth]{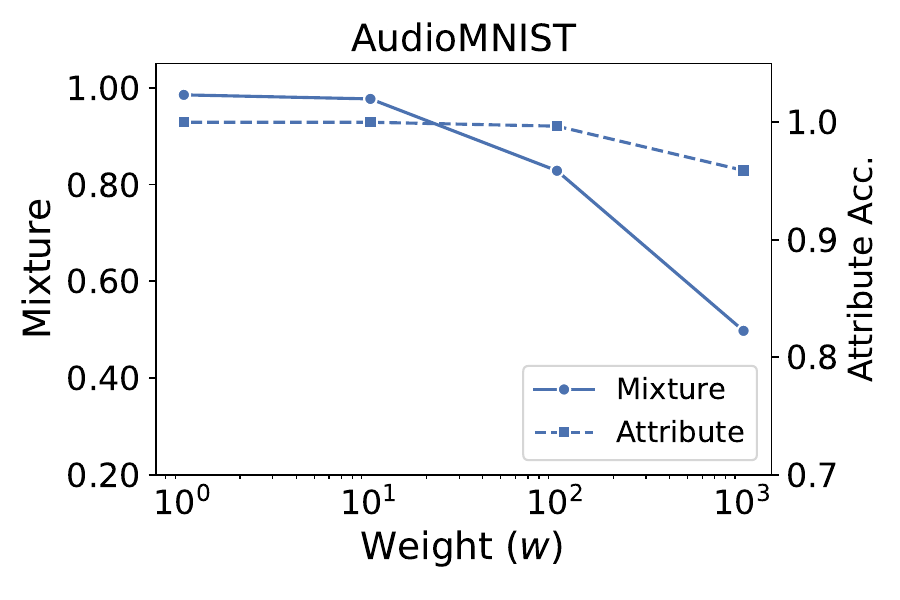}
    \caption{\footnotesize $g=128, t=0.8, r_p=50\%, r_q=0\%$}
    \end{subfigure}
    \begin{subfigure}[t]{0.31\linewidth}
        \centering
        \includegraphics[width=\linewidth]{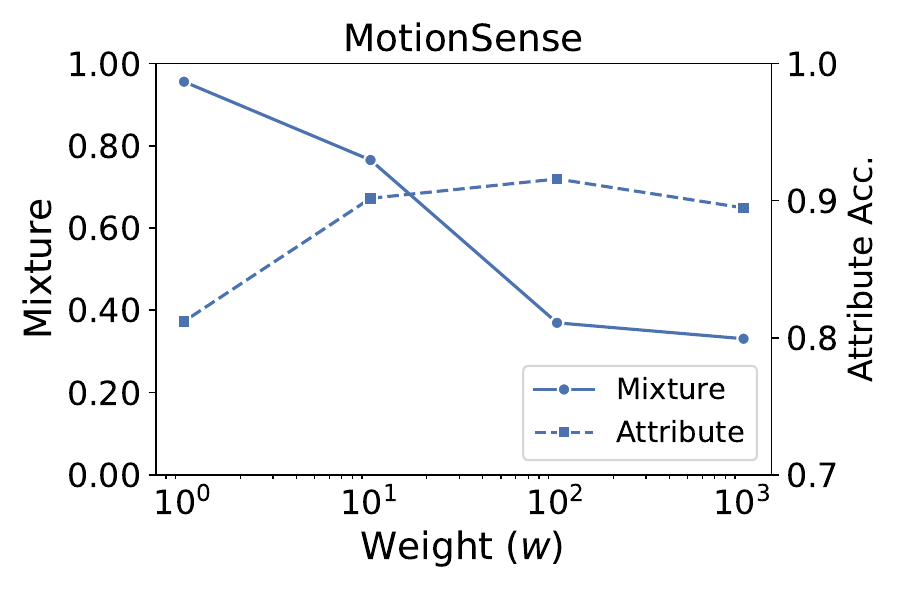}
    \caption{\highlight{\footnotesize $g=32, t=0.9, r_p=100\%, r_q=0\%$}}
    \end{subfigure}
    \caption{Varying weight $w$. Under $r_p=1\%$, our method works well regardless of the weight since only $1\%$ of features are retrained.
    On the other hand, with $r_p=50\%$, the \highlight{level of mixture} decreases when $w$ increases because the anonymized data record is now much closer to the original one because of the large portion of features being retained via a higher $r_p$ and anchored in place via a higher $w$.}
    \label{fig:abl_w}
\end{figure*}

We carried out a thorough scan through the parameter space
in order to uncover all interesting trends 
and crucial regions in our experiments.
For the sake of presentation brevity, 
we report in Sec.~\ref{sec:eval:results} only the representative results,
under the following parameter settings:
\begin{itemize}
    \item Set attribute purity $t$: ranges from $0.0$ to $1.0$ with step size $0.1$;
    \item \highlight{Set size: $g = 1, 8, 20, 32, 40, 80, 128$};
    \item Feature retention ratio $r_p = c_p / |\mathbf{d}|$ for the attribute of interest:
    $r_p = 0.1\%, 1\%, 10\%, 50\%, 100\%$;
    \item Feature retention ratio $r_q = c_q / |\mathbf{d}|$ for the additional attributes:
    $r_q=0\%, 0.1\%, 1\%, 10\%, 50\%$;
    \item Weight: \highlight{$w=1, 10, 100, 1000$},
\end{itemize}
where $c_p$, $c_q$, and $\mathbf{d}$ are as defined in Alg.~\ref{alg:alg-sampling}.
After we explored these parameters (see Fig.~\ref{fig:abl_t} through~\ref{fig:abl_p}), 
we used $t=0.6$, $g=32$, $r_p=1\%$, $w=100$ for FER-2013, $t=0.8$, $g=128$, $r_p=1\%$, $w=10$ for AudioMNIST, and \highlight{$t=0.9$, $g=32$, $r_p=10\%$, $w=100$ for MotionSense}.
\highlight{
These parameters were experimentally determined by finding the best trade-offs between classification accuracies and sensitive attribute suppressions.
}

\subsection{\highlight{Experimental} Results}
\label{sec:eval:results}
We organize our results as follows. 
First, we report the attribute recognition accuracies 
on the original unanonymized dataset as baselines.
Next, we perform ablation studies on the parameters of our method
and discuss the result.
We then take a closer examination of the quality of the anonymization
achieved by our method.
(Note that all the above results on FER-2013
are obtained by using the FaceGraph feature representation.)
Then, we experiment with 
applying our method
on all four different feature representations on FER-2013,
as discussed in Sec.~\ref{sec:eval:setup},
and report our findings.
\highlight{Moreover, as illustrated in Alg. 1, the relevance scores of features can be estimated from either the classifier or the mutual information, and our method can further incorporate the sensitive information when it is given. 
We test those two aspects in our experiments and discuss the results.
Lastly, we test our method by suppressing different sensitive information of biometric data, e.g. age and gender information.

}

\subsubsection{Performance on \highlight{the Original} Data}
Before any discussion on the anonymized biometric data,
we first establish a reference point
by obtaining the accuracy of the classification model
for the attribute of interest
on the original data.
We expect this classification result to be reasonably accurate
because otherwise it would be difficult to
assess the level of utility preservation
if the original biometric dataset 
already had low utility to begin with.
For the attributes of interest on FER-2013, AudioMNIST \highlight{and MotionSense}, 
the random-forest classifier achieved \highlight{78.8\%, 93.2\%, 89.0\%} 
recognition accuracy,
respectively.

\subsubsection{Preserving Attribute of Interest \& Suppressing Sensitive Attribute}

\begin{figure*}[t!]
    \centering
    \begin{subfigure}[t]{0.31\linewidth}
        \centering
        \includegraphics[width=\linewidth]{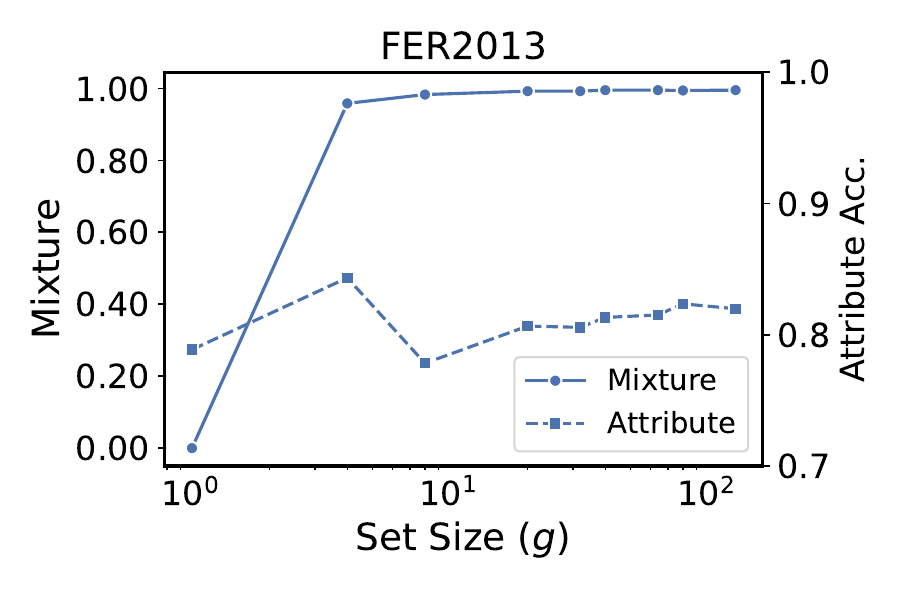}
    \caption{\footnotesize $t=0.6, w=100, r_p=1\%, r_q=0\%$}
    \end{subfigure}%
    ~ 
    \begin{subfigure}[t]{0.31\linewidth}
        \centering        
        \includegraphics[width=\linewidth]{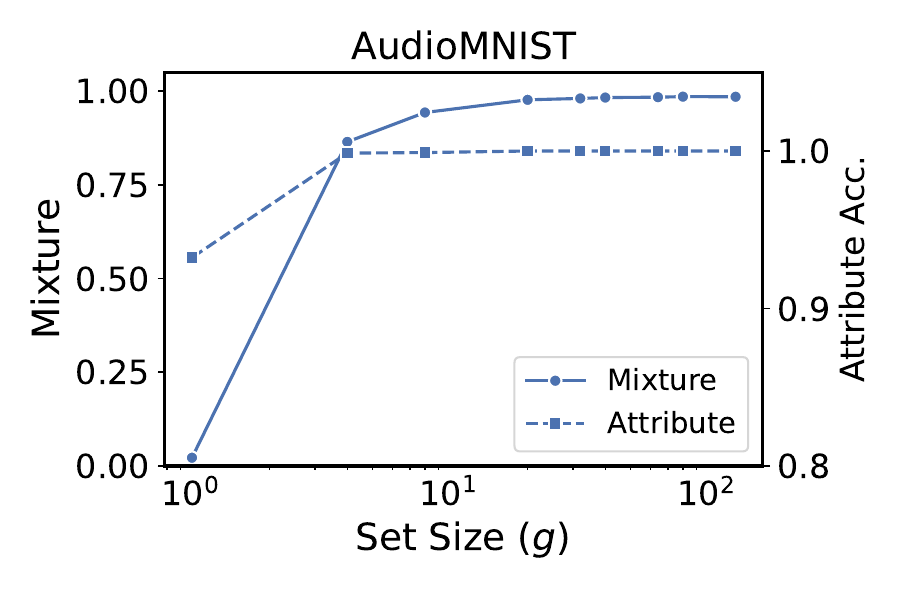}
    \caption{\footnotesize $t=0.8, w=10, r_p=1\%, r_q=0\%$}
    \end{subfigure}
    ~ 
    \begin{subfigure}[t]{0.31\linewidth}
        \centering
        \includegraphics[width=\linewidth]{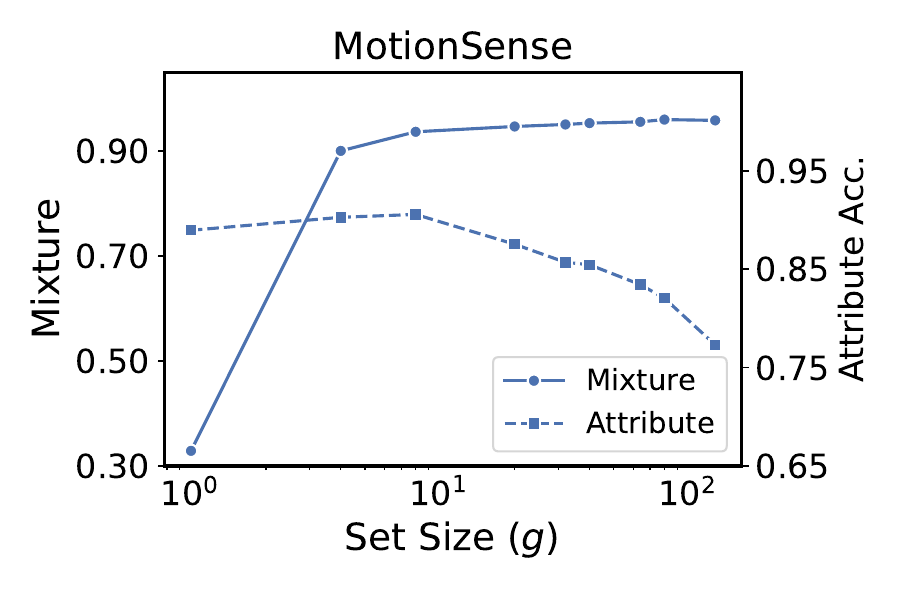}
    \caption{\highlight{\footnotesize $t=0.9, w=100, r_p=10\%, r_q=0\%$}}
    \end{subfigure}
    
    \caption{Varying set size $g$. With larger set size, \highlight{the level of mixture} increases as mixing more data leads to better anonymization without affecting the recognition of the attribute of interest.}
    \label{fig:abl_g}
\end{figure*}

\begin{figure*}[t!]
    \centering
    \begin{subfigure}[t]{0.31\linewidth}
        \centering
        \includegraphics[width=\linewidth]{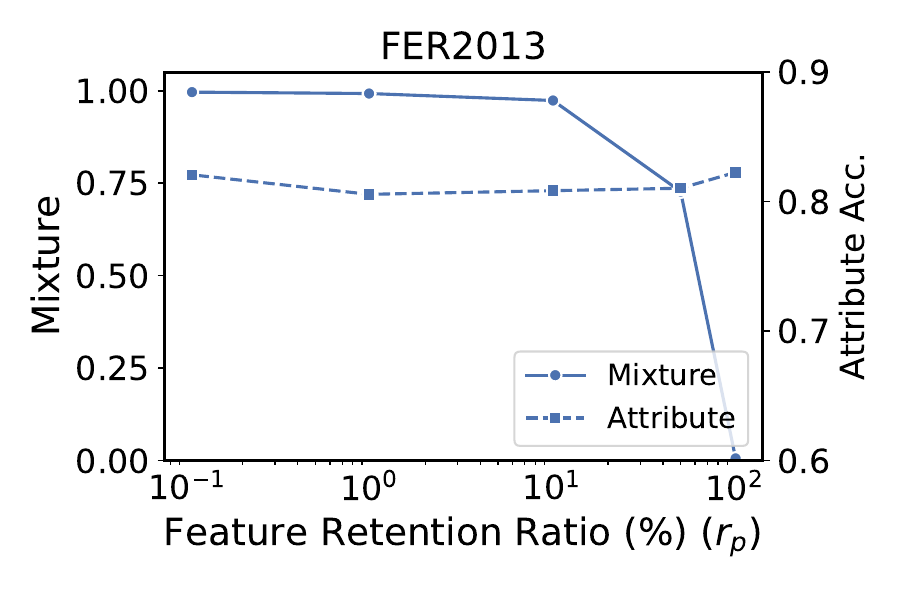}
    \caption{ $g=32, t=0.6, w=100, r_q=0\%$}
    \end{subfigure}%
    ~ 
    \begin{subfigure}[t]{0.31\linewidth}
        \centering
        \includegraphics[width=\linewidth]{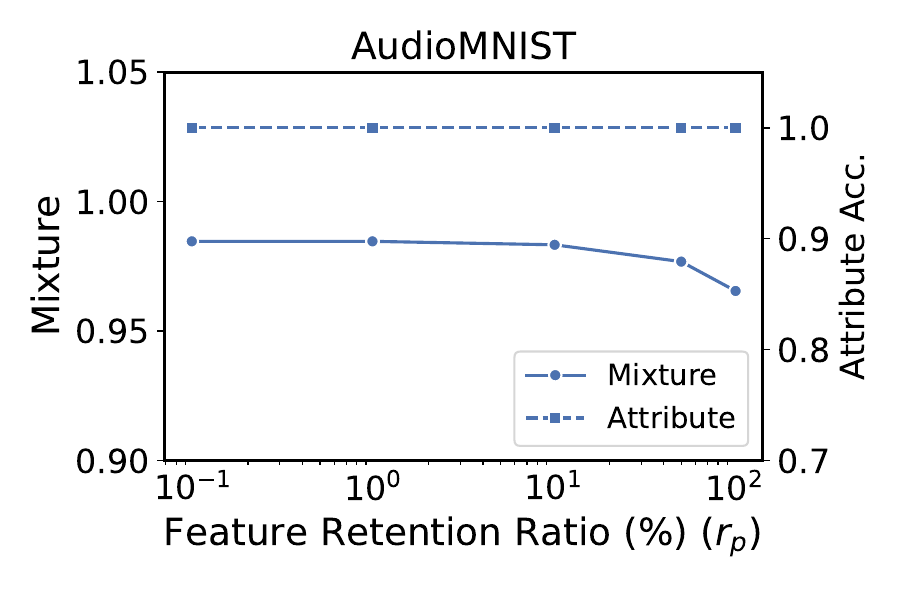}
    \caption{ $g=128, t=0.8, w=10, r_q=0\%$}
    \end{subfigure}
    ~ 
    \begin{subfigure}[t]{0.31\linewidth}
        \centering
        \includegraphics[width=\linewidth]{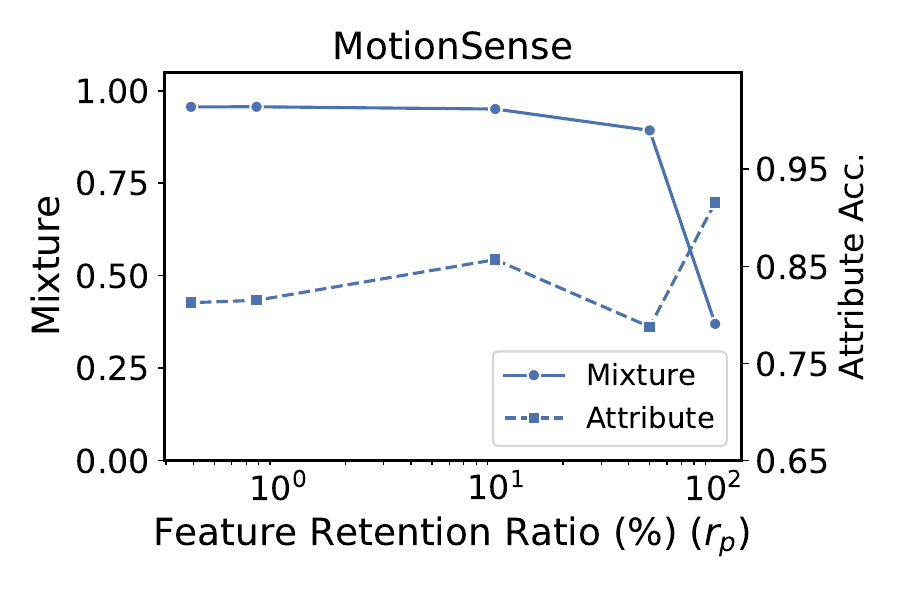}
    \caption{\highlight{ $g=32, t=0.9, w=100, r_q=0\%$}}
    \end{subfigure}
    \caption{Varying feature retention ratio $r_p$. Retaining more features increases the similarity between the original and the anonymized data,
    and can help increase attribute recognition accuracy,
    but \highlight{might lead to a lower mixture}. Hence, a trade-off 
    needs to be made here.}
    \label{fig:abl_p}
\end{figure*}

Figures~\ref{fig:abl_t} through~\ref{fig:abl_p} show how each parameter affects the data transformation's
\highlight{level of mixture} and preservation of the attribute of interest.
In each of these experiments, 
we \highlight{tuned} a single parameter while keeping the rest fixed
at the optimal configuration we obtained empirically.

First, we examined the influence of the set purity $t$, 
which determines the percentage of the data records
sharing the same attribute value as the target
in each random set,
as defined in Sec.~\ref{sec:method:clustering}.
As shown in Fig.~\ref{fig:abl_t}, 
the set purity and the recognition accuracy of the attribute of interest
on the anonymized data is positively correlated, 
which demonstrates that our method can indeed 
preserve the attribute of interest effectively.
On the other hand, 
varying the purity level 
does not affect the \highlight{level of mixture}.

The weight $w$ controls how much a data record
is anchored in place during transformation
in terms of its retained features.
Its other features would 
still be blended with the other data records.
As shown in Fig.~\ref{fig:abl_w}, 
when we set the feature retention to only keep $r_p=1\%$ of features,
even with very small weight, 
we can still achieve high recognition accuracy 
for the attribute of interest and high level of mixture
on anonymized data. 
On the other hand, when we retain $r_p=50\%$ or $100\%$ of features, 
the larger weight $w$ results in lower level of mixture 
as the anonymized data is now much too similar to the original data.

The set size $g$ is related to the size of the population
each data record is to be mixed with.
Therefore, a larger $g$ would lead to
a more diverse set for our method to increase the level of mixture. On the other hand, as we can control the set purity $t$, 
the result set will affect the \highlight{level of mixture} more 
than it does the attribute of interest. 
As shown in Fig.~\ref{fig:abl_g}, the \highlight{level of mixture} improves
when set size increases,
whereas the recognition accuracy 
of the attribute of interest remains relatively unchanged.

For the feature retention ratio $r_p$ for the attribute of interest, 
retaining more features would lead to
smaller difference between the original data record and
its anonymized version,
resulting in lower mixture,
as can be observed in Fig.~\ref{fig:abl_p}.
On the other hand, thanks to feature ranking, 
even if only $r_p=1\%$ or $10\%$ of features are retained, 
the recognition accuracy of the attribute of interest 
remains unaffected even though the \highlight{level of mixture} drops significantly.
As we expect the mixture level to be high in an anonymized dataset,
we can use such experimental parameter space exploration
to help locate desirable configuration.
For example, for the feature retention ratio in the range 
$r_p \in [0.1\%, 10\%$],
we observe, for \highlight{all three datasets},
both high levels of mixture
and high attribute recognition accuracy---both are 
desirable characteristics for utility-preserving anonymization.

\begin{figure}[t!]
    \centering
    \includegraphics[width=.31\linewidth]{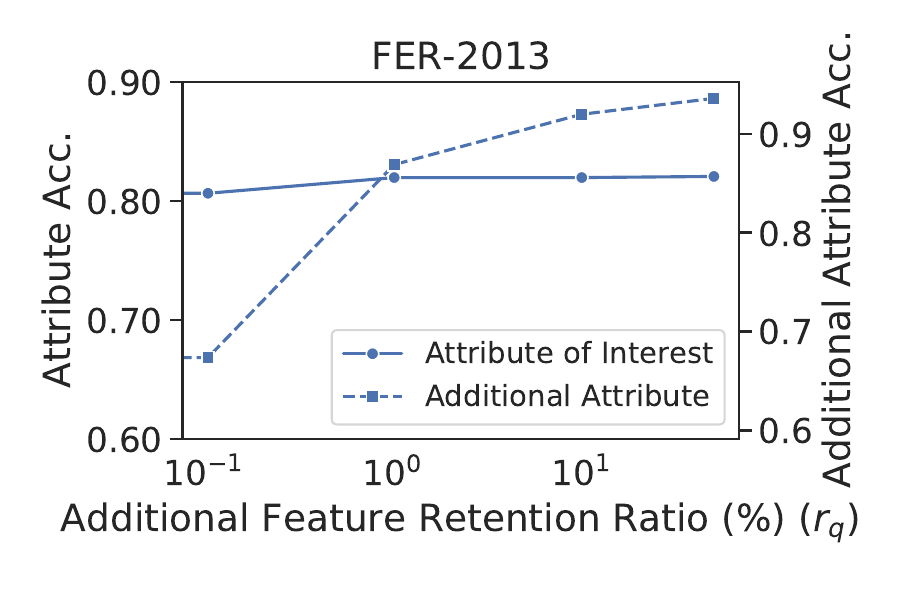}
    \caption{Varying feature retention ratio $r_q$ for the additional attribute for FER-2013 ($g=32, t=0.6, w=100, r_p=1\%$). 
    The corresponding \highlight{levels of mixture} are 0.99, 0.99, 0.97, 0.70, from left to right.
    \highlight{A trade-off} can be made here that achieves
    good recognition accuracy for both the attribute of interest
    and the additional attribute,
    as well as a high \highlight{degree of mixture}.
    }
    \label{fig:abl_s}
\end{figure}

\begin{figure*}[bt!]
    \centering
    \begin{subfigure}[t]{0.31\linewidth}
        \centering
        \includegraphics[width=\linewidth]{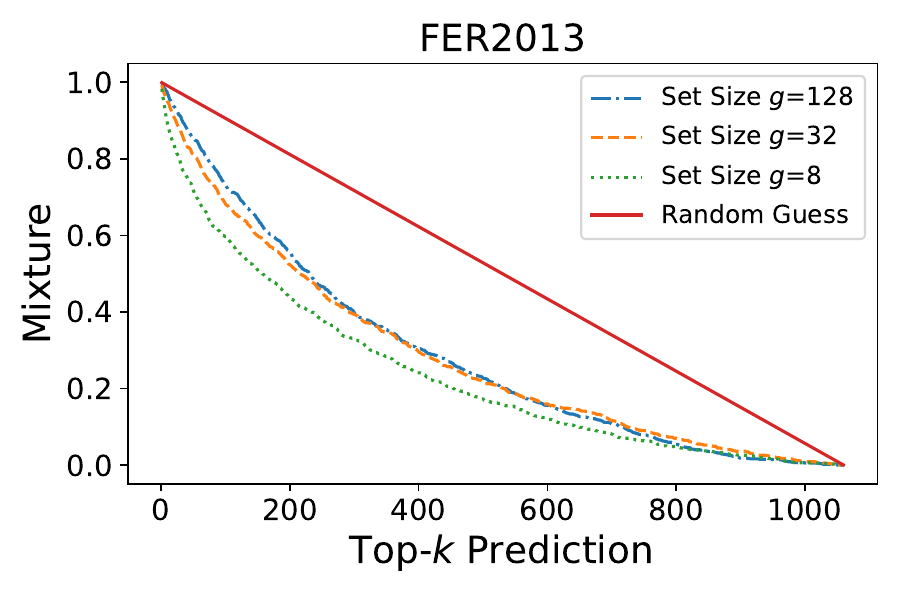}
    \caption{\footnotesize $t=0.6, w=100, r_p=1\%, r_q=0\%$}
    \end{subfigure}%
    ~ 
    \begin{subfigure}[t]{0.31\linewidth}
        \centering
        \includegraphics[width=\linewidth]{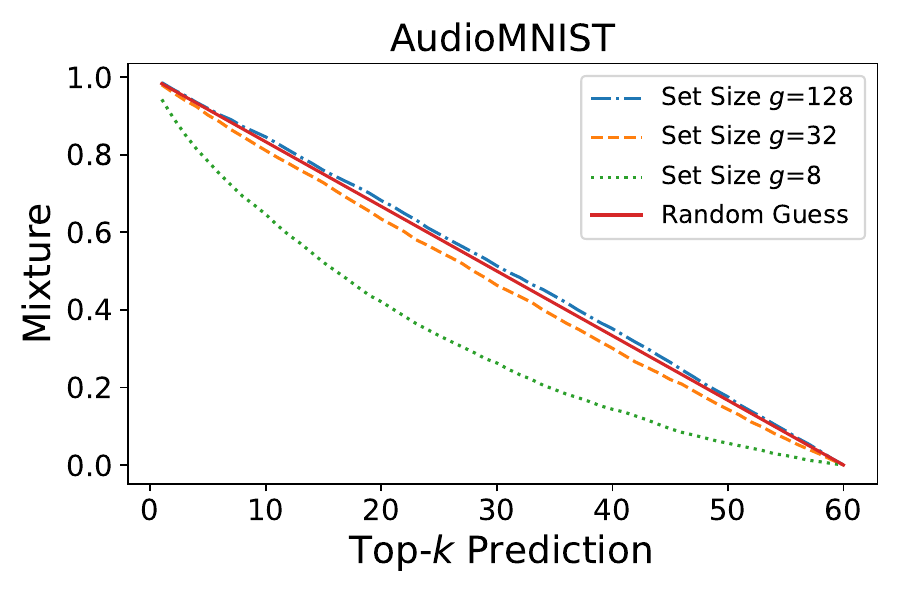}
    \caption{\footnotesize $t=0.8, w=10, r_p=1\%, r_q=0\%$}
    \end{subfigure}
    ~ 
    \begin{subfigure}[t]{0.31\linewidth}
        \centering
        \includegraphics[width=\linewidth]{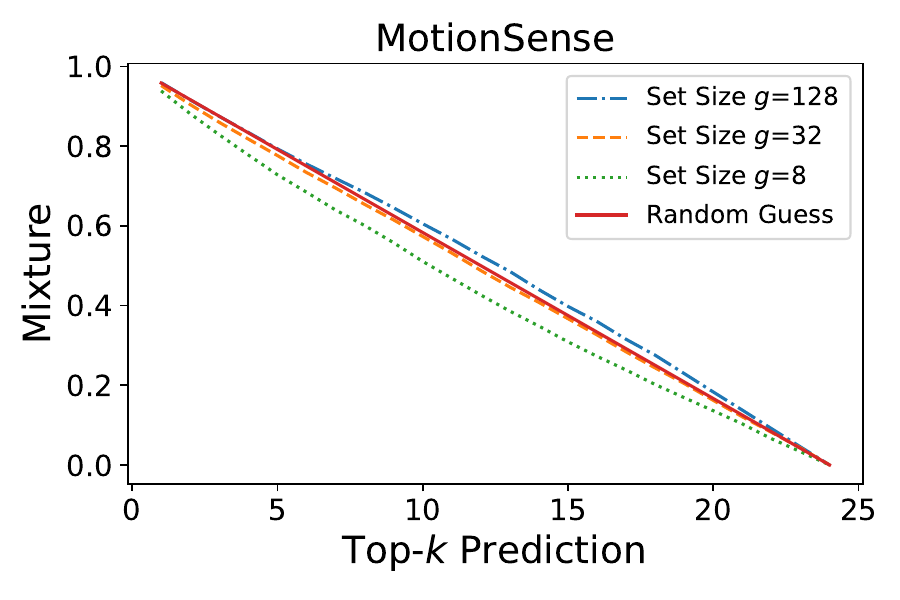}
    \caption{\highlight{\footnotesize $t=0.9, w=100, r_p=10\%, r_q=0\%$}}
    \end{subfigure}
    \caption{The \highlight{level of mixture} over top-$k$ predictions,
    where a re-identification attack 
    is considered successful if the true identity is contained in the 
    attacker's top $k$ candidate matches.
    The straight lines correspond to random guesses.
    } 
    \label{fig:top_k_vary_g}
\end{figure*}

\begin{figure*}[bt!]
    \centering
    \begin{subfigure}[t]{0.31\linewidth}
        \centering
        \includegraphics[width=\linewidth]{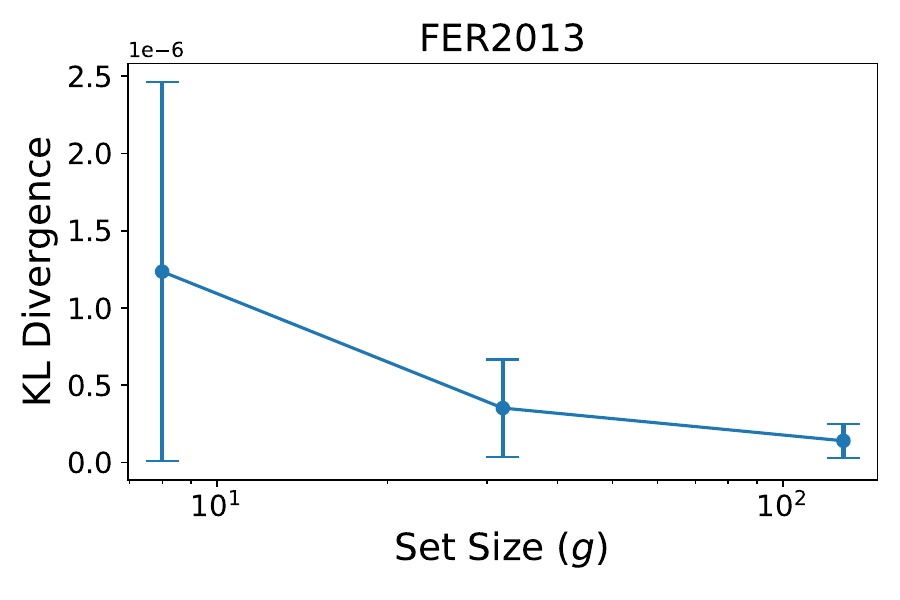}
    \caption{\footnotesize $t=0.6, w=100, r_p=1\%, r_q=0\%$}
    \end{subfigure}%
    ~ 
    \begin{subfigure}[t]{0.31\linewidth}
        \centering
        \includegraphics[width=\linewidth]{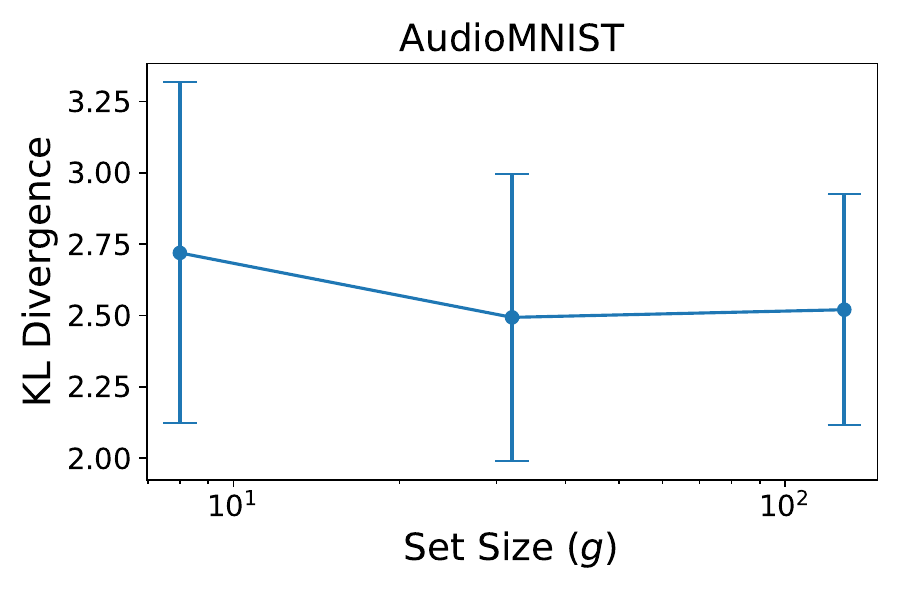}
    \caption{\footnotesize $t=0.8, w=10, r_p=1\%, r_q=0\%$}
    \end{subfigure}
    \begin{subfigure}[t]{0.31\linewidth}
        \centering
        \includegraphics[width=\linewidth]{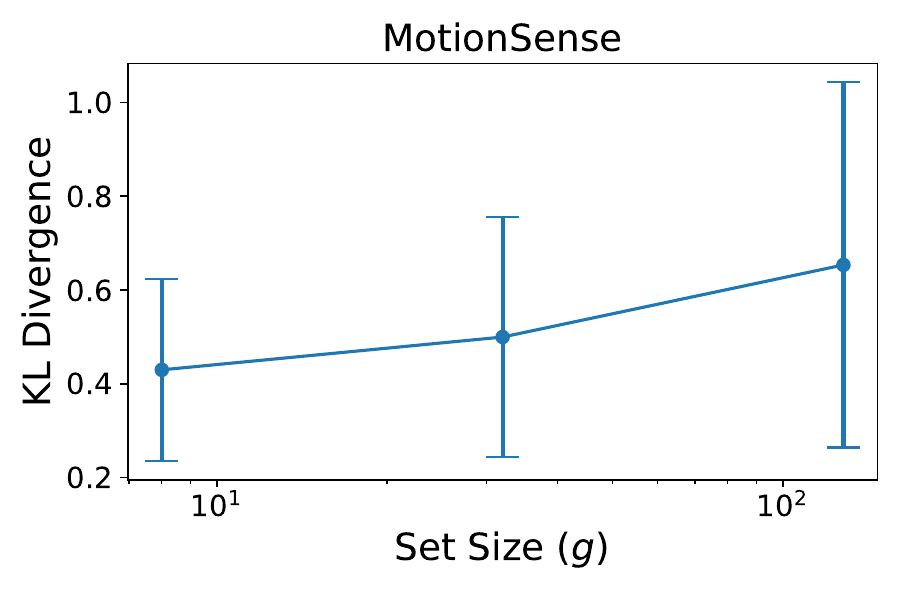}
    \caption{\highlight{\footnotesize $t=0.9, w=100, r_p=10\%, r_q=0\%$}}
    \end{subfigure}
    \caption{KL Divergence from random guess at different set size. 
    For FER-2013, the KL divergence from random guess is almost 0, whereas the overall KL divergence remains very small for AudioMNIST \highlight{and MotionSense}.}
    \label{fig:kld_vary_g}
\end{figure*}

\subsubsection{\highlight{Preserving} Additional Attribute}
We next demonstrate, using FER-2013,
the preservation of not only the attribute of interest,
but also an additional attribute,
while performing anonymization.
one of them. 
The results are shown in Fig.~\ref{fig:abl_s}.
It can be seen that when we retain more features 
related to the additional attribute, 
the recognition accuracy of the attribute of interest stays the same
while the recognition accuracy for the additional attribute
enjoys a drastic boost.
For example, when we retain just $1\%$ of 
the features for the addition attribute, 
its recognition accuracy increases by $\sim$15\% without decreasing \highlight{the mixture level},
which is at $0.99$.  
This clearly demonstrates that our method 
can effectively preserve multiple attributes when performing anonymization.

\subsubsection{Anonymization Quality} \label{sec:eval:results:quality}
We have so far been judging the quality of anonymization 
via \highlight{the level of mixture}.
While an informative metric,
the mixture level does not paint the whole picture,
as it is based only on the binary hit-or-miss results of 
identity classification models.
A ``perfect'' anonymization 
would reduce an attacker's re-identification attempts to random guesses, which means the attacker gains zero information with the attacks.
Therefore, we use two methods 
to take a deeper look into the anonymization quality
achieved by our proposed approach, with different set sizes:
{\em i)} \highlight{the level of mixture} over top-$k$ predictions, 
and 
{\em ii)} \highlight{the} KL divergence between the predicted probability 
and that of random guesses.
Results are shown in Fig.~\ref{fig:top_k_vary_g} and 
Fig.~\ref{fig:kld_vary_g}.

The \highlight{level of mixture} over top-$k$ prediction
means that a re-identification attack 
is considered successful if the true identity is contained in the 
attacker's top $k$ candidate matches.
As shown in Fig.~\ref{fig:top_k_vary_g},
if the curve is below and close to that of random guess, 
it implies that the data are anonymized in a way that 
the attacker can only achieve random guess in re-identification attacks. Moreover, if the curve is above the random guess, 
it means the anonymized data can actually fool the attacker better than random guess, 
in which case the attacker might as well try guessing randomly.

We also \highlight{measured} how far the predicted distribution deviates from
that of random guesses.
A value close to zero means that the attacker \highlight{would not} be able to
do better than random guess.
We \highlight{computed} each KL divergence from random guess 
for each data record and then \highlight{averaged} across the whole dataset.
Results are shown in Fig.~\ref{fig:kld_vary_g},
where we observe that 
{\em i)} the overall KL divergence values are already close to zero,
indicating good anonymization qualities, and 
{\em ii)} with larger set size, 
re-identification attempts tend to 
behave increasingly more like random guesses.

\subimport{./Tables/}{mi_svm}

\begin{figure*}[t!]
    \centering
    \begin{subfigure}[t]{0.31\linewidth}
        \centering
        \includegraphics[width=\linewidth]{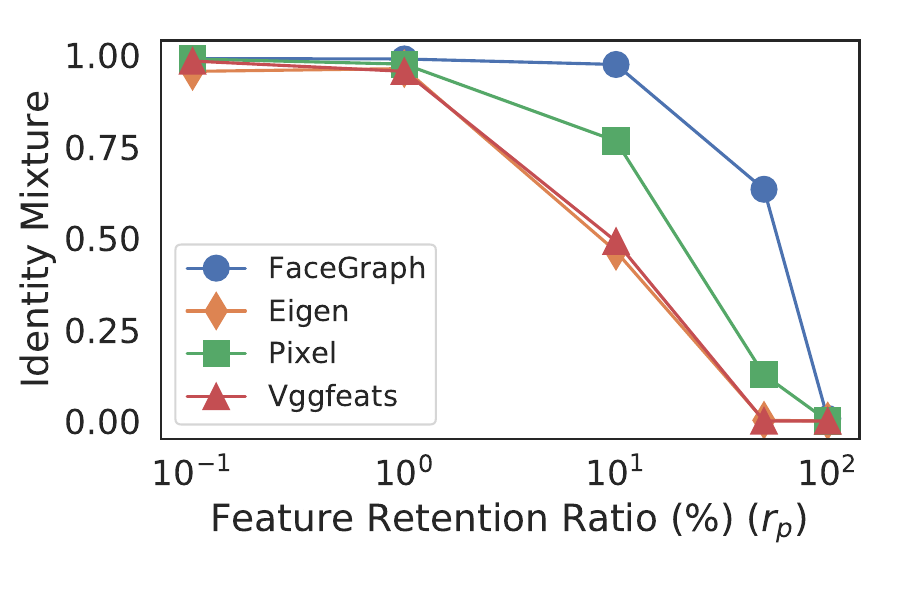}
    \end{subfigure}%
    ~ 
    \begin{subfigure}[t]{0.31\linewidth}
        \centering
        \includegraphics[width=\linewidth]{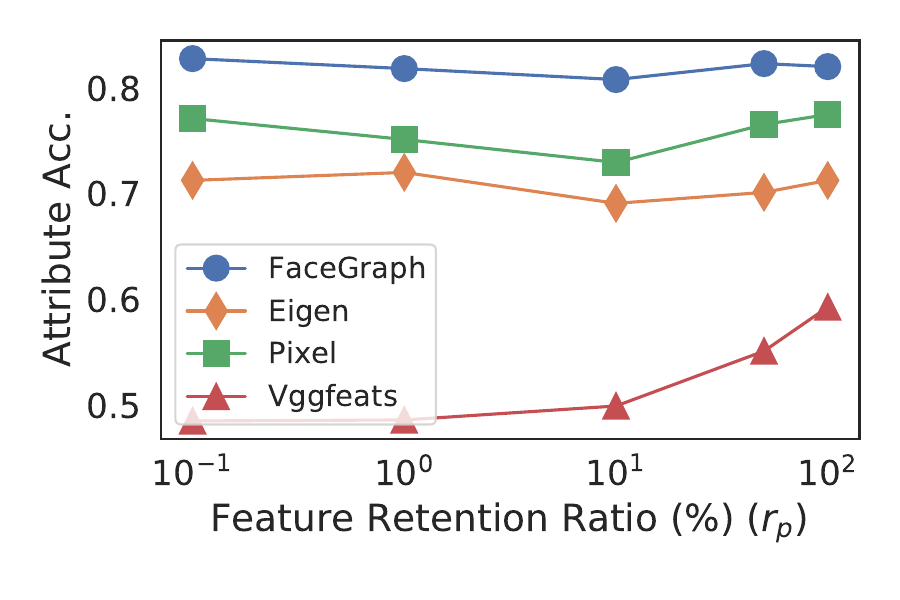}
    \end{subfigure}
    \caption{Different data representations for FER-2013. The rest of the parameter setting is $g=32, t=0.6, w=100, r_q=0\%$. Our method is applicable to different data representations, and with only 1\% features, the method can increase the mixture level to almost 1.0 while retaining good accuracy on the attribute of interest.}
    \label{fig:diff_data}
\end{figure*}

\subsubsection{Different Data Representations} %
Lastly, 
we \highlight{experimented} with multiple different biometric feature extraction \highlight{methods} and 
data representations.
Results are shown in Fig.~\ref{fig:diff_data}. 
First, it can be observed that our method is applicable to
different data representations. 
For example, when setting the feature retention ratio to $r_p\in [0.1\%, 1\%]$,
good mixture level is observed for all different data representations,
even though they do show varying recognition accuracies
for the attribute of interest.
In this particular example, 
our FaceGraph representation happens to give the best result among all. 
We also observe that the Vggfeats performs the worst,
likely due to the low resolution of the images
and the fact that the domain of images is different from that of the pretrained model.
In general, 
the optimal data representation as well as parameter configuration
can always be found by our empirical data-driven approach,
to tailor for the specific data type, utility-preserving needs,
and anonymization requirements.

\subsubsection{Different Estimators for Feature Relevance}
\label{sec:eval:diff-relevance-estimator}
\highlight{
The previous experiments all used the Random-Forest classifier,
which already privides all features' Gini importance scores.
Next, we experimentd with computing the mutual information~\cite{mutualinformation} for estimating features' relevance scores, 
which is non-parameteric and independent from any ML classifiers.
Table~\ref{tab:different_feature_ranking_and_ml_models} shows the results that use different relevance scores. 
Due to the model-agnostic nature of mutual information, 
we also experimented with other classifier such as Support Vector Machine (SVM).
Our results suggest that mutual information is also a good indicator 
for feature relevance and the selected features can perform well 
regardless of the particular choice of the classification model.
In general, using Random-Forest for both feature relevance estimation 
and the actual classification leads to good performance
as they are closely coupled within the same process.
The results from using mutual information and SVM do show slight accuracy degradation on the attribute of interest.
However, we do still observe a high level of mixture.
Therefore, our method can be extended to more diverse feature relevance estimation methods,
to be used with a wide range of different ML models.
}

\begin{figure*}[t!]
    \centering
    \begin{subfigure}[t]{0.31\linewidth}
        \centering
        \highlightfig{
        \includegraphics[width=\linewidth]{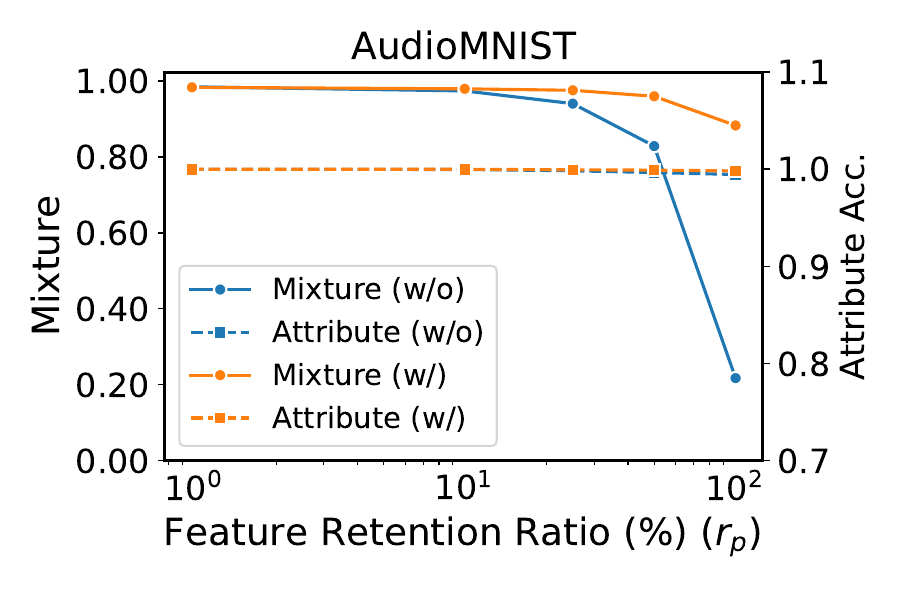}}
    \caption{$g=128, t=0.8, w=10$}
    \end{subfigure}%
    ~ 
    \begin{subfigure}[t]{0.31\linewidth}
        \centering
        \highlightfig{
        \includegraphics[width=\linewidth]{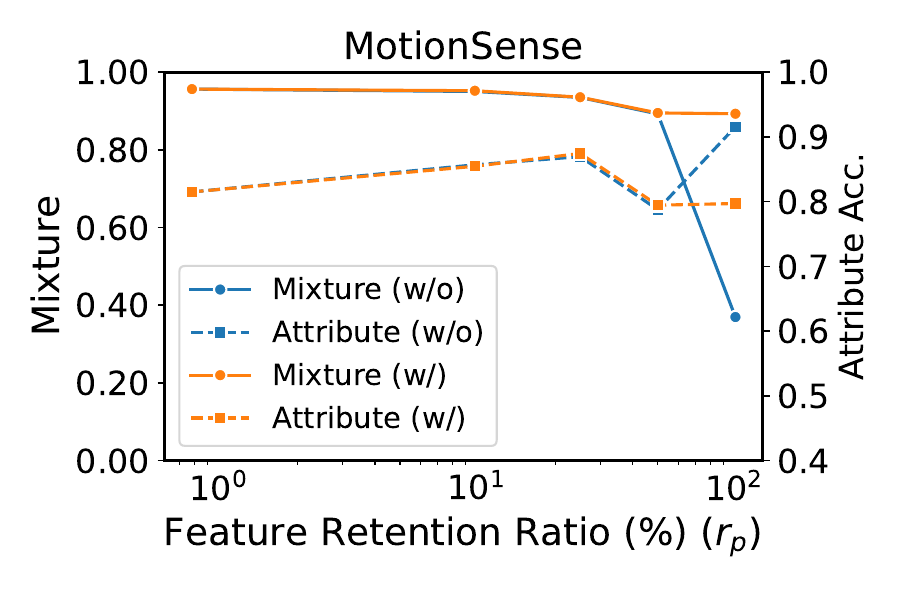}}
    \caption{$g=32, t=0.9, w=10$}
    \end{subfigure}
    \caption{\highlight{Incorporating the sensitive information into anonymization. Our method can increase the level of mixture regardless of feature retention ratio. (The $c_s$ includes top 50\% of features that are important to senstive information.)}}
    \label{fig:sensitive}
\end{figure*}

\subsubsection{Incorporating Sensitive Attribute}
\label{subsubsec:suppress}
\highlight{
So far in our experiments,
we have been selecting features only based on their relevance to 
the attributes we want to preserve.
However, as discussed in Sec.~\ref{sec:method:weighted-mean},
and shown in Alg.~\ref{alg:alg-sampling}, 
our method can also incorporate feature relevance to 
the sensitive attribute that we are aiming to suppress from the original data.
In this experiment, 
we take the ranked feature list built for the attribute of interest
and remove from it the features that are important to the sensitive attribute. 
That is, $\{c_p\} \setminus \{c_s\}$, where $\{c_p\}$ and $\{c_s\}$ are the set of selected features based on the feature retention ratio. 
As FER2013 does not include the sensitive information explicitly, 
we only perform this experiment on AudioMNIST and MotionSense.
We use the Random-Forest classifier for the sensitive attribute and directly use the provided feature importance scores as the relevance measure.
Figure~\ref{fig:sensitive} demonstrates the effectivness of including the sensitive information into the anonymization process.
For AudioMNIST, as the feature retention ratio increases, 
the level of mixture drops drastically when feature selection was done without accounting for the sensitive attribute, 
leading to poor anonymization quality.
However, on the other hand, 
the level of mixture remains high as our method rejected the features that are important to the sensitive attribute. 
A similiar trend can also be observed for the experiment on MotionSense.
}

\subsubsection{\highlight{Suppressing Other Sensitive Attribute}}\label{subsubsec:other_sensitive}
\subimport{./Tables/}{other_sensitive_attribute}

\highlight{
In addition to suppressing the identity information in the previous sets of experiments, 
we also test the flexiblity of our method by suppressing other sensitive attributes.
Specifically, the \textit{age} attribute in AudioMNIST and the \textit{gender} attribute in MotionSense. 
Table~\ref{tab:other_sensitive_attributes} shows the results of the level of mixture and the attribute of interest on two datasets.
Our method achieves good accuracy on the attribute of interest while reducing the attempt to recover the sensitive attribute to random guess. This shows that our method can be easily configured to handle different type of sensitive attributes.
}

%% file: Tables/mi_svm.tex
\begin{table*}[t]
    \centering
    \caption{\highlight{Experimenting with different feature relevance estimations and machine learning models. RF: Random Forest; MI: Mutual Information; SVM: Support Vector Machine. As shown, our proposed method generalizes well to different combinations of feature ranking methodsand machine learning models.}}
    \label{tab:different_feature_ranking_and_ml_models}
    \begin{adjustbox}{max width=\linewidth}
\highlight{
    \begin{tabular}{cc|cc|cc|cc}
    \toprule
        &   &   \multicolumn{2}{c|}{FER2013} & \multicolumn{2}{c|}{AudioMNIST} & \multicolumn{2}{c}{MotionSense} \\
        Feature & ML &  \multirow{2}{*}{Mixture}  & Attribute &  \multirow{2}{*}{Mixture}  & Attribute &  \multirow{2}{*}{Mixture}  & Attribute  \\
       Ranking & Model &  & of Interest  & & of Interest & & of Interest  \\
       \midrule
    RF & RF & 0.99 & 0.81 & 0.98 & 1.0 & 0.95 & 0.86\\
    MI & RF & 0.99 & 0.84 & 0.98 & 1.0 & 0.95 & 0.84\\
    MI & SVM & 0.99 & 0.79 & 0.98 & 1.0  & 0.95 & 0.73\\
    \bottomrule
    \multicolumn{8}{l}{\scriptsize The hyper-parameters ($t$, $w$, $g$ and $r_p$) are selected with the best performance as described in Sec.~\ref{subsubsec:parameters}.}
    \end{tabular}
}
    \end{adjustbox}
\end{table*}

%% file: Tables/other_sensitive_attribute.tex
\begin{table}[t]
    \centering
    \caption{\highlight{Suppression of different sensitive attributes (other than identity). 
    In this set of experiments, AudioMNIST's \textit{age} and MotionSense's \textit{gender}
    are considered to be the sensitive attributes.
    Their levels of mixtures are greatly increased on the anonymized data,
    whereas the accuracies for their respective attributes of interest remain largely unaffected.
    which demonstrates our algorithm's flexible in suppressing different types of sensitive attibutes.}}
    \label{tab:other_sensitive_attributes}
    \begin{adjustbox}{max width=\linewidth}
\highlight{
    \begin{tabular}{c|cc|cc}
    \toprule
        Dataset & \multicolumn{2}{c|}{AudioMNIST}  &  \multicolumn{2}{c}{MotionSense} \\  
         & Attribute of & Mixture & Attribute of & Mixture \\
         & Interest Accuracy & (Age) & Interest Accuracy & (Gender) \\
         \midrule
        Original Data & 0.93 & 0.35   & 0.89 & 0.17  \\
        Anonymized Data  & 1.0 & 0.84  & 0.86 & 0.49 \\
    \bottomrule
    \multicolumn{5}{l}{\scriptsize The hyper-parameters ($t$, $w$, $g$ and $r_p$) are selected with the best performance as described in Sec.~\ref{subsubsec:parameters}.}
    \end{tabular}
}
    \end{adjustbox}
\end{table}

%% file: Tex/02_related_works.tex
\section{Related Work}\label{sec:related_work}

\highlight{Closely related work can be considered those
that apply or adapt the data truthfulness-preserving anonymity techniques,
such as $k$-anonymity~\cite{samarati2001protecting},
$\ell$-diversity~\cite{Machanavajjhala06}, 
and $t$-closeness~\cite{Li07},
to various types of data,}
ranging from categorical that might appear in relational database tables, 
to location and biometric data.
Typically, most applications seek to find that balance between anonymizing the data effectively 
while also retaining utility to some degree~\cite{dfls.ijufks2012, cdfs.kadmds2007}. 
The fundamental difference of our proposed method from these existing techniques 
is that \textit{we do not generalize}, as each transformed biometric data record
remains different from the others,
which opens up the possibility of more interesting attributes being preserved
through anonymization.
\highlight{
Although categorical data was one of the first sources for applications of $k$-anonymity, $\ell$-diversity, and $t$-closeness,
which our proposed method was inspired by,
applications to biometric data (e.g., facial images) is the most relevant to the contribution of this paper.
For example,
the $k$-same family of algorithms~\cite{newton2005preserving, gross2005integrating, Gross08, Sun15}
were proposed to achieve anonymization of facial images while retraining utility.
With slight variations, these algorithms essentially
partition a facial image dataset to clusters of size $k$
and replace the $k$ individuals in a cluster with their corresponding centroid.
In this way every individual in the cluster shares the same de-identified face (i.e., the centroid). 
The $k$-same family of algorithms were not extended to enforce $\ell$-diversity in attributes 
that could be considered sensitive. 
Furthermore, many of the instantiations of $k$-same operate in pixel-space, 
leading to degradation in the utility of the anonymity representations,
e.g., via excessive blurring induced by the centroid computation. 
Other related work, such as $k$-Diff-furthest~\cite{Sun15}, extend $k$-same to allow distinguishable representations for the anonymized data, so that for example, individuals can be tracked in videos without revealing their identity. 
However, $k$-Diff-furthest only focuses on the distinguishability among anonymized individuals,
neglecting their truthfullness and utility for downstream machine learning-based analytical tasks.
}

\highlight{

Broadly speaking,
privacy protection techniques generally fall into two categories.
The first notion of privacy is the protection against \textit{membership inference attacks},
which dictates that 
queries on the entire dataset produce approximately the same results
after small perturbation to the dataset.
A widely studied technique for protecting this notion of privacy
is \textit{Differential Privacy}~\cite{TCS-042},
which has also been applied to the facial recognition setting.
For example, PEEP (Privacy using EigEnface Perturbation)~\cite{CHAMIKARA2020101951}
is a local differential privacy method 
that improves the robustness of trained models against membership inference attacks.
The other notion of privacy is \textit{information-theoretic privacy}~\cite{hsu2021survey},
which refers to privacy protections that also retain the conditional distribution
of query results on individual data samples.
Our proposed method in this paper falls under this second type of privacy protection.
}

Recent works have explored more advanced anonymization models such as neural networks 
for face de-identification~\cite{Meden17,pan2019ksamesiamesegan,Li_2019_CVPR_Workshops}. 
While showing impressive clarity in generating fake faces for replacing real faces, 
these techniques require large amounts of training data and are also difficult to interpret or reason about, 
making it difficult to audit the models for industrial applications. 
It is also in question as to whether the focus should be on accurate reproduction of life-like anonymized images in pixel-space 
or a focus on generating highly anonymized abstract representations that can retain utility for other tasks, 
we follow the latter approach in this paper. 
The survey article~\cite{Ribaric16} 
\highlight{discusses additional
related work 
on face de-identification}.

\highlight{
A first version of our work appeared in ESORICS 2022~\cite{moriarty2022utility}. 
The work here is considerably extended with enhanced algorithm designs 
as well as expanded experimental evaluation.
More specifically, as opposed to solely relying on the classification method 
to provide features' task relevance measures,
we also employ model-agnostic, purely data-driven statistical models
(e.g., mutual information)
to estimate feature relevance scores.
Additionally, when determining the subset of features to select for weighting,
we now also take into consideration their relevance to the identity 
or any other sensitive attributes that we want to suppress,
rather than only considering their relevance to attributes we intend to preserve.
Furthermore, we expand our evaluation with a more complete set of abalation studies,
and include a new dataset, MotionSense~\cite{Malekzadeh:2019:MSD:3302505.3310068},
to our experiments, besides facial images and voice audios.
Our thorough experimental evaluation using three completely different data modalities 
demonstrates the promising performance as well as the flexibility and generalizability
of our proposed method.
}

%% file: Tex/05_conclusions.tex
\section{Conclusion}\label{sec:conclusions}
In this paper we introduce
a biometric data transformation framework
that aims at stripping away \highlight{sensitive (e.g., personally identifiable) information}
while at the same time preserving the utility of the biometrics
by leaving \highlight{its other characteristics intact}
such that downstream machine learning-based analytical tasks
could still extract useful and valuable attributes from the \highlight{transformed} biometric data.
We present our \highlight{novel end-to-end data transformation} algorithm design,
which, \highlight{for each biometric record,
first forms a \facegroup,
then uses either classification model-based or purely data-driven statistical approaches 
to estimate its features' relevance to all the valuable as well as sensitive attributes,
and finally, using the feature relevance scores,
carries out a selective weighted-mean-based data transformation.}
We experimentally evaluated our method
using publicly available facial \highlight{imagery, voice audio, and human activity motion} datasets
and observed that our proposed method
could effectively \highlight{suppress sensitive attributes} from the different modalities of biometrics,
while at the same time successfully preserve other interesting attributes for downstream machine learning-based analytics.

%% file: main.bbl
\begin{thebibliography}{10}

\bibitem{swift-vision}
Apple.
\newblock Vision framework: Apply computer vision algorithms to perform a
  variety of tasks on input images and video.
\newblock \url{https://developer.apple.com/documentation/vision}, 2021.

\bibitem{bdgps.access2021}
M.~Barni, R.~{Donida Labati}, A.~Genovese, V.~Piuri, and F.~Scotti.
\newblock Iris deidentification with high visual realism for privacy protection
  on websites and social networks.
\newblock {\em IEEE Access}, 9:131995--132010, 2021.
\newblock 2169-3536.

\bibitem{audiomnist}
S\"oren Becker, Marcel Ackermann, Sebastian Lapuschkin, Klaus-Robert M\"uller,
  and Wojciech Samek.
\newblock Interpreting and explaining deep neural networks for classification
  of audio signals.
\newblock {\em CoRR}, abs/1807.03418, 2018.

\bibitem{breiman2001random}
Leo Breiman.
\newblock Random forests.
\newblock {\em Machine learning}, 2001.

\bibitem{CHAMIKARA2020101951}
M.A.P. Chamikara, P.~Bertok, I.~Khalil, D.~Liu, and S.~Camtepe.
\newblock Privacy preserving face recognition utilizing differential privacy.
\newblock {\em Computers \& Security}, 97:101951, 2020.

\bibitem{cdfs.kadmds2007}
V.~Ciriani, S.~{De Capitani di Vimercati}, S.~Foresti, and P.~Samarati.
\newblock {k-Anonymity}.
\newblock In T.~Yu and S.~Jajodia, editors, {\em Secure Data Management in
  Decentralized Systems}. Springer-Verlag, 2007.

\bibitem{datta2020survey}
Priyanka Datta, Shanu Bhardwaj, Surya~Narayan Panda, Sarvesh Tanwar, and Sumit
  Badotra.
\newblock Survey of security and privacy issues on biometric system.
\newblock In {\em Handbook of Computer Networks and Cyber Security}, pages
  763--776. Springer, 2020.

\bibitem{dfls.ijufks2012}
S.~{De Capitani di Vimercati}, S.~Foresti, G.~Livraga, and P.~Samarati.
\newblock Data privacy: Definitions and techniques.
\newblock {\em International Journal of Uncertainty, Fuzziness and
  Knowledge-Based Systems}, 20(6):793--817, December 2012.

\bibitem{dpd.springer2013}
R.~{Donida Labati}, V.~Piuri, and F.~Scotti.
\newblock Biometric privacy protection: guidelines and technologies.
\newblock In M.~S. Obaidat, J.S. Sevillano, and F.~Joaquim, editors, {\em
  Communications in Computer and Information Science}, volume 314, pages 3--19.
  Springer, 2012.
\newblock 978-3-642-35754-1.

\bibitem{TCS-042}
Cynthia Dwork and Aaron Roth.
\newblock The algorithmic foundations of differential privacy.
\newblock {\em Foundations and Trends® in Theoretical Computer Science},
  9(3–4):211--407, 2014.

\bibitem{goodfellow2013challenges}
Ian~J Goodfellow, Dumitru Erhan, Pierre~Luc Carrier, Aaron Courville, Mehdi
  Mirza, Ben Hamner, Will Cukierski, Yichuan Tang, David Thaler, Dong-Hyun Lee,
  et~al.
\newblock Challenges in representation learning: A report on three machine
  learning contests.
\newblock In {\em NIPS}, 2013.

\bibitem{gross2005integrating}
Ralph Gross, Edoardo Airoldi, Bradley Malin, and Latanya Sweeney.
\newblock Integrating utility into face de-identification.
\newblock In {\em PET}, 2005.

\bibitem{Gross08}
Ralph Gross, Latanya Sweeney, Fernando de~la Torre, and Simon Baker.
\newblock Semi-supervised learning of multi-factor models for face
  de-identification.
\newblock In {\em CVPR}, 2008.

\bibitem{FisherFeatureSelection}
Quanquan Gu, Zhenhui Li, and Jiawei Han.
\newblock Generalized fisher score for feature selection.
\newblock In {\em Proceedings of the Twenty-Seventh Conference on Uncertainty
  in Artificial Intelligence}, UAI'11, page 266–273, Arlington, Virginia,
  USA, 2011. AUAI Press.

\bibitem{hsu2021survey}
Hsiang Hsu, Natalia Martinez, Martin Bertran, Guillermo Sapiro, and Flavio~P
  Calmon.
\newblock A survey on statistical, information, and estimation—theoretic
  views on privacy.
\newblock {\em IEEE BITS the Information Theory Magazine}, 1(1):45--56, 2021.

\bibitem{Hubert}
Wei-Ning Hsu, Benjamin Bolte, Yao-Hung~Hubert Tsai, Kushal Lakhotia, Ruslan
  Salakhutdinov, and Abdelrahman Mohamed.
\newblock Hubert: Self-supervised speech representation learning by masked
  prediction of hidden units.
\newblock {\em IEEE/ACM Transactions on Audio, Speech, and Language
  Processing}, 29:3451--3460, 2021.

\bibitem{Li07}
Ninghui Li, Tiancheng Li, and Suresh Venkatasubramanian.
\newblock $t$-closeness: Privacy beyond $k$-anonymity and $\ell$-diversity.
\newblock In {\em ICDE}, 2007.

\bibitem{Li_2019_CVPR_Workshops}
Tao Li and Lei Lin.
\newblock Anonymousnet: Natural face de-identification with measurable privacy.
\newblock In {\em CVPR Workshops}, 2019.

\bibitem{liu2015faceattributes}
Ziwei Liu, Ping Luo, Xiaogang Wang, and Xiaoou Tang.
\newblock Deep learning face attributes in the wild.
\newblock In {\em ICCV}, 2015.

\bibitem{Machanavajjhala06}
Ashwin Machanavajjhala, Daniel Kifer, Johannes Gehrke, and Muthuramakrishnan
  Venkitasubramaniam.
\newblock $\ell$-diversity: Privacy beyond $k$-anonymity.
\newblock {\em ACM TKDD}, 2007.

\bibitem{Malekzadeh:2019:MSD:3302505.3310068}
Mohammad Malekzadeh, Richard~G. Clegg, Andrea Cavallaro, and Hamed Haddadi.
\newblock Mobile sensor data anonymization.
\newblock In {\em Proceedings of the International Conference on Internet of
  Things Design and Implementation}, IoTDI '19, pages 49--58, New York, NY,
  USA, 2019. ACM.

\bibitem{Meden17}
Blaz Meden, Ziga Emersic, Vitomir Struc, and Peter Peer.
\newblock $\kappa$-same-net: neural-network-based face deidentification.
\newblock In {\em IWOBI}, 2017.

\bibitem{moriarty2022utility}
Bill Moriarty, Chun-Fu Chen, Shaohan Hu, Sean Moran, Marco Pistoia, Vincenzo
  Piuri, and Pierangela Samarati.
\newblock Utility-preserving biometric information anonymization.
\newblock In {\em European Symposium on Research in Computer Security}, pages
  24--41. Springer, 2022.

\bibitem{newton2005preserving}
Elaine~M Newton, Latanya Sweeney, and Bradley Malin.
\newblock Preserving privacy by de-identifying face images.
\newblock {\em IEEE TKDE}, 2005.

\bibitem{ortiz2018survey}
Nicolas Ortiz, Ruben~Dario Hern{\'a}ndez, Robinson Jimenez, Mauricio
  Mauledeoux, and Oscar Avil{\'e}s.
\newblock Survey of biometric pattern recognition via machine learning
  techniques.
\newblock {\em Contemporary Engineering Sciences}, 11(34):1677--1694, 2018.

\bibitem{pan2019ksamesiamesegan}
Yi-Lun Pan, Min-Jhih Haung, Kuo-Teng Ding, Ja-Ling Wu, and Jyh-Shing Jang.
\newblock K-same-siamese-gan: K-same algorithm with generative adversarial
  network for facial image de-identification with hyperparameter tuning and
  mixed precision training.
\newblock In {\em AVSS}, 2019.

\bibitem{scikit-learn}
F.~Pedregosa, G.~Varoquaux, A.~Gramfort, V.~Michel, B.~Thirion, O.~Grisel,
  M.~Blondel, P.~Prettenhofer, R.~Weiss, V.~Dubourg, J.~Vanderplas, A.~Passos,
  D.~Cournapeau, M.~Brucher, M.~Perrot, and E.~Duchesnay.
\newblock Scikit-learn: Machine learning in {P}ython.
\newblock {\em Journal of Machine Learning Research}, 2011.

\bibitem{Ribaric16}
Slobodan Ribaric, Aladdin Ariyaeeinia, and Nikola Pavesic.
\newblock De-identification for privacy protection in multimedia content.
\newblock {\em Image Commun.}, 2016.

\bibitem{InfiniteFeatureSelection}
Giorgio Roffo, Simone Melzi, and Marco Cristani.
\newblock Infinite feature selection.
\newblock In {\em 2015 IEEE International Conference on Computer Vision
  (ICCV)}, pages 4202--4210, 2015.

\bibitem{mutualinformation}
Brian~C. Ross.
\newblock Mutual information between discrete and continuous data sets.
\newblock {\em PLOS ONE}, 9(2):1--5, 02 2014.

\bibitem{rui2018survey}
Zhang Rui and Zheng Yan.
\newblock A survey on biometric authentication: Toward secure and
  privacy-preserving identification.
\newblock {\em IEEE access}, 7:5994--6009, 2018.

\bibitem{samarati2001protecting}
Pierangela Samarati.
\newblock {Protecting Respondents' Identities in Microdata Release}.
\newblock {\em IEEE Transactions on Knowledge and Data Engineering (TKDE)},
  13(6):1010--1027, November/December 2001.

\bibitem{FaceNet}
Florian Schroff, Dmitry Kalenichenko, and James Philbin.
\newblock Facenet: A unified embedding for face recognition and clustering.
\newblock In {\em Proceedings of the IEEE Conference on Computer Vision and
  Pattern Recognition (CVPR)}, June 2015.

\bibitem{Sun15}
Zongji Sun, Li~Meng, and Aladdin Ariyaeeinia.
\newblock Distinguishable de-identified faces.
\newblock In {\em FG}, 2015.

\end{thebibliography}
